\documentclass{article}
\PassOptionsToPackage{numbers, compress}{natbib}
\usepackage[dblblindworkshop, final]{neurips_2026}

\usepackage[utf8]{inputenc} 
\usepackage[T1]{fontenc}    
\usepackage{hyperref}       
\hypersetup{hidelinks}
\usepackage{url}            
\usepackage{booktabs}       
\usepackage{amsfonts}       
\usepackage{nicefrac}       
\usepackage{microtype}      
\usepackage{enumitem}
\usepackage{multirow}
\usepackage{makecell}
\usepackage{etoc}
\usepackage{marvosym}

\usepackage{graphicx}
\graphicspath{{Figs/}}
\usepackage{amsmath}
\usepackage{amssymb} 
\usepackage{xcolor}
\usepackage{colortbl}
\usepackage{booktabs}
\usepackage{xspace}
\usepackage{tikz}
\usepackage{pgfplots}
\pgfplotsset{compat=1.18}
\usepackage{newtxtt} 
\usepackage{inconsolata}
\usepackage[type1,scaled=1]{biolinum}
\usepackage{amssymb}
\usepackage{pifont}

\newcommand{\alg}{%
  \mbox{\raisebox{-0.25em}{\includegraphics[height=1em]{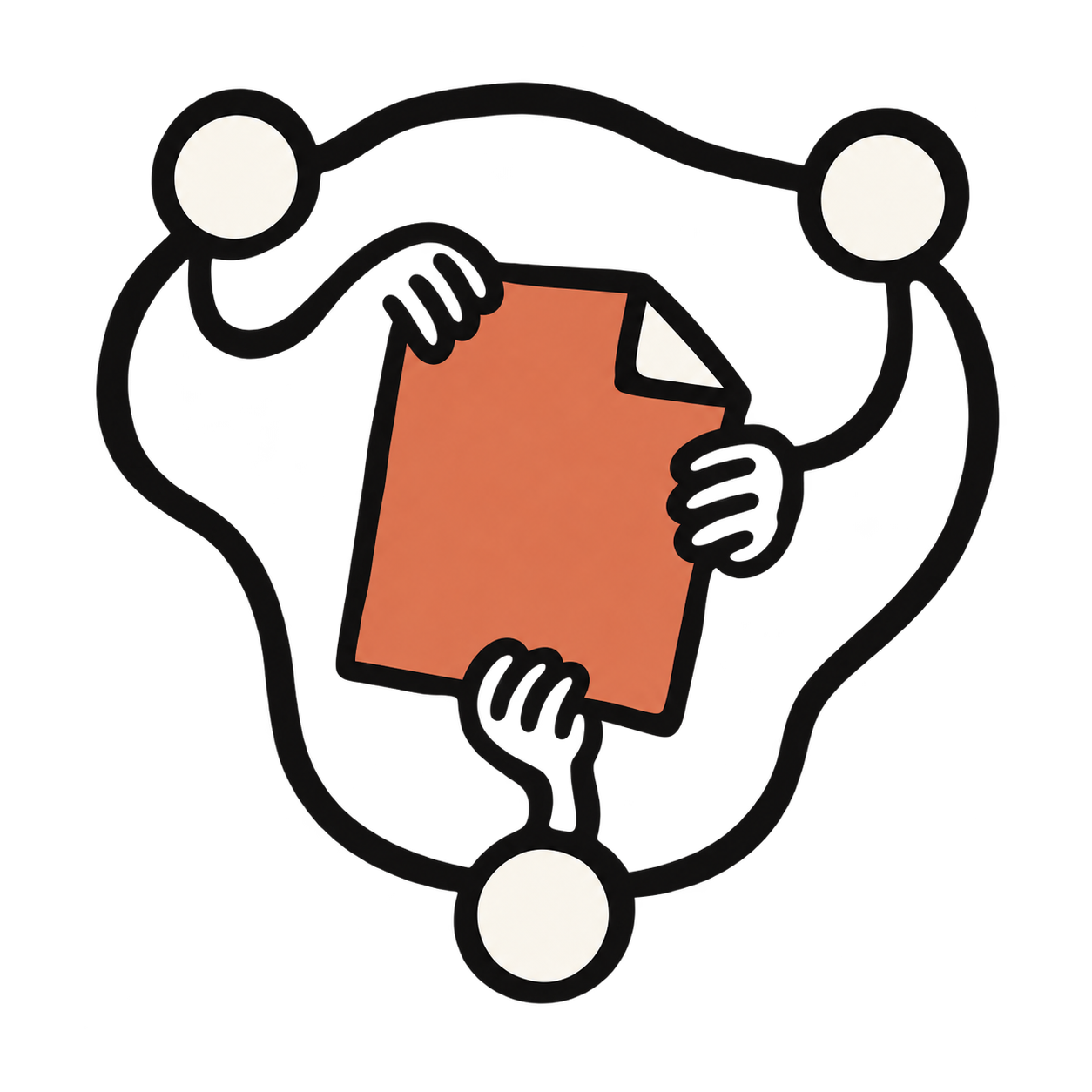}}\kern1.2pt\texttt{CABAL}}%
  \xspace
}

\usepackage{tabularx}
\usepackage{listings}
\definecolor{artifactbg}{HTML}{F7F8FA}
\definecolor{artifactframe}{HTML}{D8DEE8}
\lstdefinestyle{artifactjson}{
    basicstyle=\ttfamily\fontsize{7.2}{8.4}\selectfont,
    columns=fullflexible,
    keepspaces=true,
    breaklines=true,
    breakatwhitespace=false,
    showstringspaces=false,
    frame=single,
    framesep=4pt,
    rulecolor=\color{artifactframe},
    backgroundcolor=\color{artifactbg},
    xleftmargin=0.25em,
    xrightmargin=0.25em,
    aboveskip=0.35em,
    belowskip=0.15em
}

\definecolor{grayblue}{RGB}{232,239,246}
\title{%
  \parbox[c]{2.5em}{\raisebox{-0.1em}{\includegraphics[height=2.3em]{Figs/logo.png}}}\;
  \parbox[c]{\dimexpr\textwidth-2.8em\relax}{
    \texttt{CABAL}: Multi-Agent Simulacra for Tracing the \\
    Effects of Collusive Bidding in Peer Review
  }%
}

\workshoptitle{AI-Native Academia: Authorship, Peer Review, and Conference Governance under AI}

\author{%
\bfseries
Jicheng Zhou$^{1}$,
Kemou Li$^{1}$,
Kahim Wong$^{1}$,
Zheyuan Li$^{1}$,
\\
\bfseries
Zhuan Shi$^{2, 3}$,
Fengpeng Li$^{4}$,
Haiwei Wu$^{5}$,
Jiantao Zhou$^{1}$\textsuperscript{\Letter}
\\[4pt]
\mdseries
$^{1}$State Key Laboratory of Internet of Things for Smart City, University of Macau
\\
$^{2}$Mila -- Québec AI Institute \qquad $^{3}$McGill University
\\
$^{4}$PRADA Lab, King Abdullah University of Science and Technology
\\
$^{5}$University of Electronic Science and Technology of China%
}

\begin{document}
\etocdepthtag.toc{mtchapter}
\maketitle


\begingroup
\renewcommand{\thefootnote}{\Letter}
\footnotetext[1]{%
  \begin{tabular}[t]{@{}l@{}}
    Corresponding author: Jiantao Zhou
    (\texttt{jtzhou@um.edu.mo}). \\
    Code is available at: \url{https://github.com/RupertZJC/CABAL}.
  \end{tabular}%
}
\endgroup
\setcounter{footnote}{0}

\begin{abstract}
    Recent reports during the AAAI-27 review cycle highlight the risk of reviewers coordinating bids for reciprocal assignment advantage. 
    Prior work treats bidding, reviewer assignment, and review manipulation as separate stages, leaving the lifecycle effects of collusive bidding unclear. Real-world analysis is further constrained by typically unobservable collusive intent and the lack of counterfactuals for the same conference.
    Motivated by this gap, we introduce \alg, an end-to-end multi-agent simulacra framework for studying reviewer assignment integrity by holding the conference environment fixed and configuring LLM-driven reviewer agents with honest or collusive policies. We further develop an affinity-guided collusive bidding strategy that uses mutual reviewer-paper affinities to construct collusion rings and select target papers, producing expertise-consistent rather than arbitrarily targeted attacks. 
    Controlled experiments show that collusive bidding more than doubles target-paper capture and that assigned colluders score target papers about two points higher than honest co-reviewers, while conference-wide effects remain comparatively modest. Evaluated bid-phase detectors provide only limited evidence of collusion: in a fixed-triplet detector stress test, native positive-bid graphs are confounded by benign affinity, while a Very-High-only diagnostic view enables precise but low-coverage local recovery.
\end{abstract}

\section{Introduction}
\label{sec:intro}
\begin{figure}[tbp]
      \centering
      \includegraphics[width=1.0\linewidth]{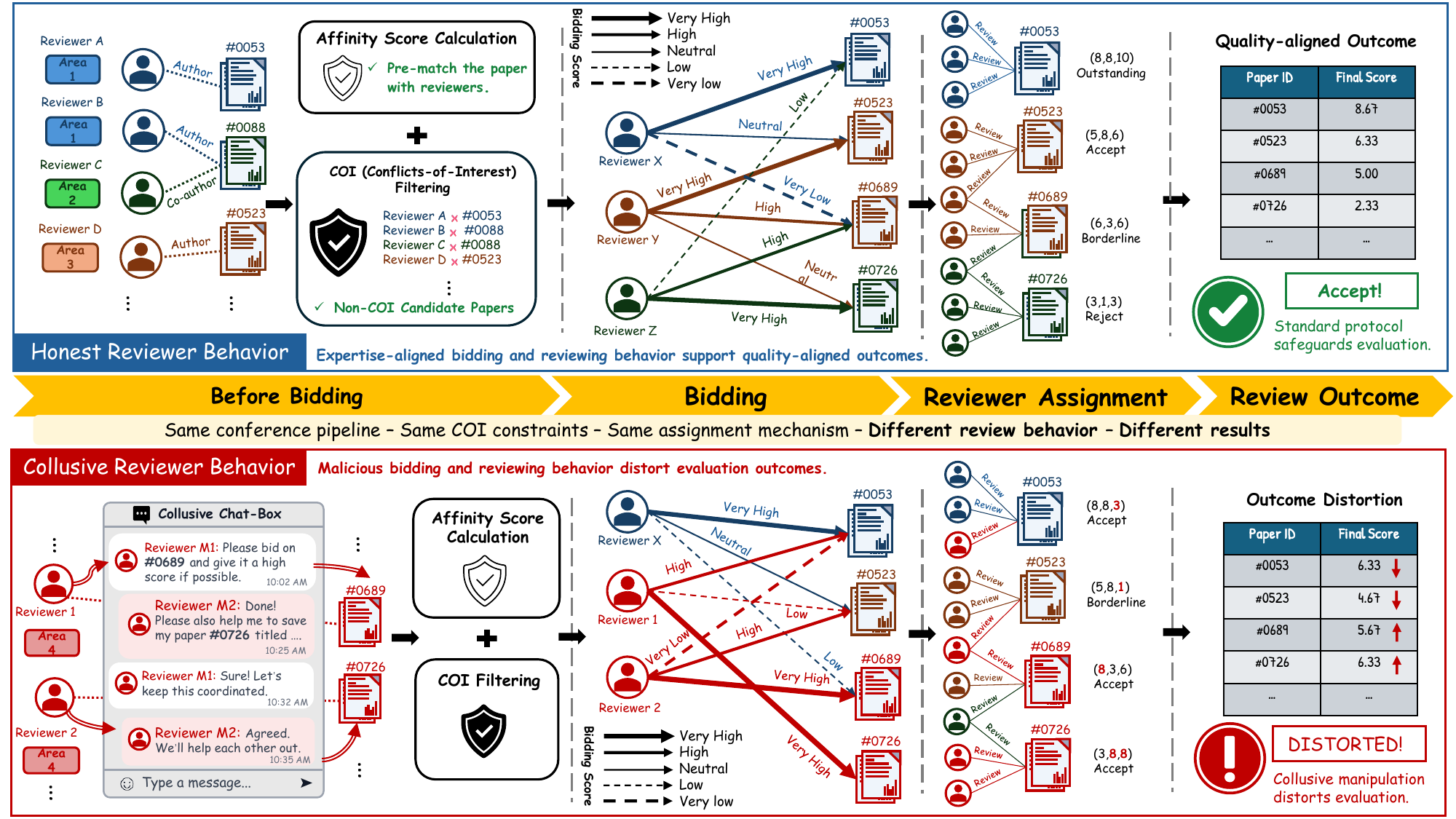}
      \caption{\textbf{Overview of the lifecycle effects studied in \alg.} With the conference environment (papers, reviewer profiles, COI constraints, and assignment mechanism) held fixed, honest reviewer behavior is expertise-aligned, whereas collusive reviewer behavior involves coordinated, affinity-guided bids to capture target-paper assignments and distort downstream evaluations.}
      \label{fig:fig1}
    \end{figure}

    In top-tier conferences, reviewers bid on papers to express their interests, and organizers combine these bids with reviewer-paper affinity, conflict-of-interest (COI) constraints, and workload requirements to determine assignments. However, this reliance on bidding also creates an attack surface for strategic manipulation. This risk became concrete during the AAAI-27 review cycle, when AAAI reported evidence suggesting off-platform coordination intended to create reciprocal assignment structures~\cite{aaai2026xreports}. In response, AAAI introduced constraints against reciprocal 2-cycles and an auditing protocol for residual cases, while warning that confirmed violations could result in desk rejection or multi-year bans~\cite{aaai2026biddingweb}. Fig.~\ref{fig:fig1} illustrates this pathway: honest bidding supports expertise-aligned assignments, whereas coordinated bids can capture assignments and distort downstream evaluations.

    Prior work has examined reviewer matching and assignment optimization \cite{mimno2007expertise,charlin2012framework,goldsmith2007ai,kobren2019paper,stelmakh2021peerreview4all,fiez2020super,rozencweig2023mitigating}, strategic bid manipulation and reviewer collusion \cite{jecmen2020mitigating,wu2021making,jecmen2023dataset,boehmer2022combating,jecmen2025detection,hsieh2025vulnerability}, and LLM-based peer-review agents \cite{jin2024agentreview,lu2025agent}. While these lines of work address complementary aspects of the review process, they do not jointly connect collusive bidding, reviewer assignment, and downstream evaluation within a controlled setting. As a result, how expertise-grounded collusive bids propagate through assignment and ultimately affect review outcomes remains insufficiently understood.
    
    Studying these lifecycle effects in real conferences poses two key challenges. 
    First, collusive intent is typically unobserved, making expertise-consistent collusive bids difficult to distinguish from legitimate expressions of reviewer interest. 
    Second, real conferences lack the causal counterfactual needed to isolate the effect of reviewer behavior: the same conference cannot be rerun with its papers, reviewers, COI constraints, and assignment mechanism held fixed while changing only whether reviewers behave collusively. As a result, observed differences in review outcomes cannot be cleanly attributed to collusive behavior itself. Addressing these challenges therefore requires a controlled environment in which collusive intent is known, and all other conference conditions are held fixed across behavioral worlds.


    We introduce \textbf{\alg} (\textbf{C}ollusive \textbf{A}gent-based \textbf{B}idding and \textbf{A}ssignment \textbf{L}aboratory), an end-to-end multi-agent simulacra framework for controlled studies of how collusive bidding affects reviewer assignment integrity and downstream evaluation. 
    Within a fixed conference environment, LLM-driven reviewer simulacra follow known honest or collusive policies. An affinity-guided bidding strategy derives rings and target papers from mutual reviewer-paper affinities, preserving plausible expertise matches. Tab.~\ref{tab:related_comparison} contrasts this joint treatment of collusion, bidding, assignment, and reviewing with representative prior work.
    Our contributions are summarized as follows:  
    \begin{itemize}
    [
    leftmargin=*,
    itemsep=0.4em,
    topsep=0.3em,
    parsep=0em,
    partopsep=0em]

        \item We introduce \alg, a programmable multi-agent simulacra framework for matched counterfactual comparisons of bidding, assignment, and reviewing, holding papers, reviewers, COI constraints, and the assignment mechanism fixed while varying reviewer behavior.
        

        \item We develop an affinity-guided collusive bidding strategy that constructs rings and selects target papers from mutual reviewer-paper affinities, enabling reciprocal support on papers that plausibly match each colluder's expertise.
        

        \item We conduct an end-to-end empirical analysis of how collusive bids alter targeted reviewer-paper access, propagate into downstream review scores and conference-level outcomes, and compare the resulting attack patterns against representative bid-phase detection methods.
    \end{itemize}

\begin{table*}[t]
    \centering
    \setlength{\tabcolsep}{3.2pt}

    \caption{
    Comparison with representative prior work across the peer-review lifecycle.
    \alg jointly connects expertise-grounded strategic behavior, bidding,
    assignment, downstream reviewing, and outcome analysis within a controlled
    behavioral counterfactual.
    }
    \label{tab:related_comparison}

    \resizebox{\textwidth}{!}{%
    \begin{tabular}{llccccccc}
        \toprule

        \small\makecell{Work}
        & \small\makecell{Venue}
        & \small\makecell{Expertise/\\Affinity}
        & \small\makecell{Strategic\\Behavior}
        & \small\makecell{Bidding}
        & \small\makecell{Assignment}
        & \small\makecell{Reviewing}
        & \small\makecell{Outcome \\Analysis}
        & \small\makecell{Behavioral\\Counterfactual}
        \\

        \midrule

        \rowcolor{grayblue}
        \multicolumn{9}{l}{
        \textbf{Reviewer Assignment and Bidding Mechanisms}
        } \\

        \addlinespace[2pt]

        Anjum et al.~\cite{anjum2019pare}
        & \textit{EMNLP-IJCNLP'19}
        & $\bullet$ & - & -
        & $\circ$ & - & - & -
        \\

        Stelmakh et al.~\cite{stelmakh2021peerreview4all}
        & \textit{JMLR'21}
        & $\circ$ & - & -
        & $\bullet$ & - & $\circ$ & -
        \\

        Rozencweig et al.~\cite{rozencweig2023mitigating}
        & \textit{AAMAS'23}
        & $\circ$ & - & $\bullet$
        & $\circ$ & - & - & $\circ$
        \\

        \addlinespace[4pt]

        \rowcolor{grayblue}
        \multicolumn{9}{l}{
        \textbf{Strategic Manipulation and Adversarial Behaviors in Peer Review}
        } \\

        \addlinespace[2pt]

        Wu et al.~\cite{wu2021making}
        & \textit{ICML'21}
        & $\circ$ & $\bullet$ & $\bullet$
        & $\bullet$ & - & - & $\circ$
        \\

        Jecmen et al.~\cite{jecmen2023dataset}
        & \textit{WWW'23}
        & $\circ$ & $\bullet$ & $\bullet$
        & $\bullet$ & - & - & $\circ$
        \\

        Jecmen et al.~\cite{jecmen2025detection}
        & \textit{TMLR'25}
        & $\circ$ & $\bullet$ & $\bullet$
        & - & - & - & $\circ$
        \\

        Hsieh et al.~\cite{hsieh2025vulnerability}
        & \textit{USENIX Sec'25}
        & $\bullet$ & $\bullet$ & -
        & $\bullet$ & - & - & $\circ$
        \\

        \addlinespace[4pt]

        \rowcolor{grayblue}
        \multicolumn{9}{l}{
        \textbf{LLM-based Peer-Review Agents and Simulation}
        } \\

        \addlinespace[2pt]

        Jin et al.~\cite{jin2024agentreview}
        & \textit{EMNLP'24}
        & $\circ$ & - & -
        & - & $\bullet$ & $\bullet$ & $\bullet$
        \\

        Lu et al.~\cite{lu2025agent}
        & \textit{ICML'25}
        & $\bullet$
        & -
        & -
        & -
        & $\bullet$
        & $\circ$
        & -
        \\

        \addlinespace[3pt]
        \midrule

        \textbf{\alg}
        & \textit{\textbf{Ours}}
        & $\bullet$
        & $\bullet$
        & $\bullet$
        & $\bullet$
        & $\bullet$
        & $\bullet$
        & $\bullet$
        \\

        \bottomrule
    \end{tabular}%
    }

    \vspace{2pt}

    \parbox{\textwidth}{%
        \scriptsize
        \emph{Notes.}
        $\bullet$ indicates that a component is explicitly modeled or is a
        primary object of study;
        $\circ$ indicates that it is incorporated as an input, auxiliary
        component, or downstream analysis;
        and - indicates that it is not modeled.
    }
\end{table*}

\section{Problem Statement} 
\label{sec:formulation} 
    \textbf{Problem formulation.}
    We consider a conference with papers
    $\mathcal{P}=\{p_1,\ldots,p_N\}$ and reviewers
    $\mathcal{R}=\{r_1,\ldots,r_M\}$. For each paper-reviewer pair
    $(p,r_j)$, $A_{pj}$ denotes their topical affinity score, with a larger
    value indicating a stronger expertise match, and $C_{pj}\in\{0,1\}$
    denotes the conflict-of-interest (COI) indicator, where $C_{pj}=1$
    means that reviewer $r_j$ is ineligible to review paper $p$.
    We compare two behavioral worlds indexed by $z\in\{h,c\}$, where $h$
    denotes the all-honest world and $c$ denotes the collusive world. 
    Given the bid matrix $\mathbf{B}^{(z)}$, affinity matrix $\mathbf{A}$,
    and COI matrix $\mathbf{C}$, a fixed assignment procedure $\Phi$
    produces the binary assignment matrix $\mathbf{M}^{(z)}$, where
    $M_{pj}^{(z)}=1$ indicates that reviewer $r_j$ is assigned to paper
    $p$. For each assigned pair, $Y_{pj}^{(z)}$ denotes the numerical
    recommendation score given by reviewer $r_j$ to paper $p$. The review
    lifecycle is therefore
    \begin{equation}
    \begin{aligned}
        \mathbf{M}^{(z)}
        &= \Phi\!\left(
            \mathbf{B}^{(z)},\mathbf{A},\mathbf{C}
        \right), \\
        \mathbf{B}^{(z)}
        &\longrightarrow
        \mathbf{M}^{(z)}
        \longrightarrow
        \mathbf{Y}^{(z)},
        \qquad z\in\{h,c\}.
    \end{aligned}
    \label{eq:review_lifecycle}
    \end{equation}

    \textbf{Threat model.} Our threat model considers groups of reviewer-authors with
    overlapping expertise who can legitimately bid on and may be assigned
    to one another's submissions. Such expertise overlap is a practical
    condition for plausible collusion, because reviewers outside a
    paper's relevant research area are typically less viable candidates
    for its assignment. We define collusive bidding as off-platform
    coordination through which these reviewer-authors strategically
    manipulate their bidding preferences to increase the likelihood of
    reciprocal assignments. Colluders operate through the ordinary
    bidding and reviewing interfaces: they neither control the assignment
    mechanism nor bypass its COI and eligibility constraints. If
    coordinated bidding results in the intended assignments, they may
    subsequently evaluate one another's papers strategically rather than
    independently. This threat model therefore captures coordinated
    behavior across bidding, assignment, and reviewing without prescribing
    how colluding groups are formed, how bids are manipulated, or how
    strategic reviews are generated.

    \textbf{Reviewer assignment integrity.} We use reviewer assignment integrity to denote the resistance of assignment outcomes to coordinated manipulation of reviewer bids.
    Integrity is compromised when such coordination gives colluders
    greater access to one another's submissions than they would obtain
    under honest behavior. We examine this effect first at the assignment
    level by asking whether designated reviewers obtain their intended
    papers, and then at the review level by asking whether successful
    assignment capture alters their evaluations. Assignment capture is
    therefore the direct integrity effect, whereas review distortion is
    its downstream consequence.
    
    \textbf{Counterfactual analysis.} Our analysis follows a matched counterfactual design. Each collusive world is paired with an all-honest world that shares the same papers,
    reviewers, affinity structure, COI constraints, and assignment
    procedure; only reviewer behavior varies. This comparison isolates
    whether changes in bidding propagate into reviewer assignments,
    target-paper evaluations, and conference-wide outcomes. We
    additionally evaluate whether representative bid-phase detectors can
    distinguish the resulting collusive behavior from honest bidding.

\section{\texorpdfstring{\alg}{CABAL}: Collusive Agent-based Bidding and Assignment Laboratory}
\begin{figure}[tbp]
    \centering
    \includegraphics[width=1.0\linewidth]{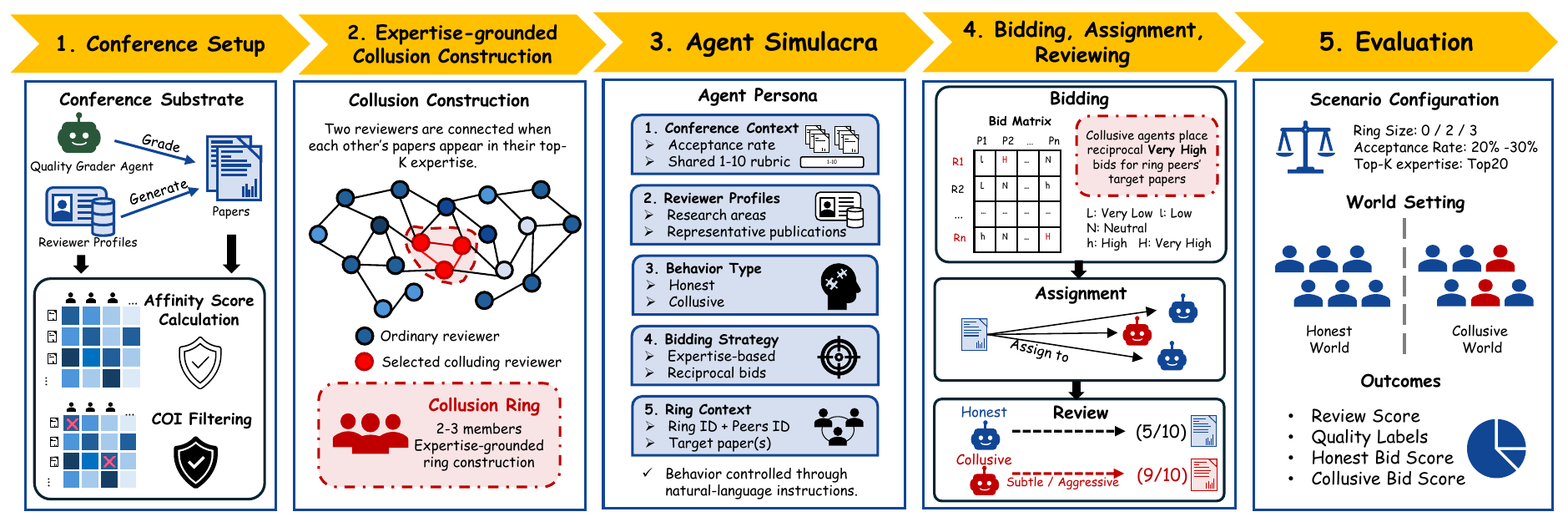}
    \caption{\textbf{Overview of \alg.} The workflow comprises five stages: (1) constructing a fixed conference substrate; (2) forming expertise-grounded collusion rings and target papers; (3) instantiating honest or collusive reviewer simulacra; (4) executing bidding, assignment, and reviewing; and (5) evaluating outcomes across matched behavioral worlds.}
    \label{fig:fig2}
\end{figure}

    Fig.~\ref{fig:fig2} provides an overview of \alg, which operationalizes the peer-review lifecycle in Eq.~\eqref{eq:review_lifecycle} as a controlled multi-agent workflow. \alg first constructs a fixed conference substrate shared across behavioral worlds (\S\ref{sec:environment}), then derives expertise-grounded collusion rings and target papers from mutual reviewer-paper affinity (\S\ref{sec:collusion}) and instantiates reviewer simulacra with honest or collusive behavioral objectives (\S\ref{sec:simulacra}). These agents proceed through bidding, assignment, and reviewing to produce $\mathbf{B}^{(z)}$, $\mathbf{M}^{(z)}$, and $\mathbf{Y}^{(z)}$ (\S\ref{sec:pathway}), enabling matched comparisons that isolate how reviewer behavior propagates through assignment into downstream evaluation.
    
\subsection{Conference Substrate} 
\label{sec:environment} 
    \alg instantiates the fixed conference substrate defined in \S\ref{sec:formulation}. Public researcher profiles, including research areas and representative publications, define the reviewer pool $\mathcal{R}$ and ground reviewer expertise. Synthetic submissions $\mathcal{P}$ are generated from author backgrounds, and an independent grading agent assigns reference-quality scores using the shared rubric without access to reviewer behavior or collusion context.

    Let $\mathcal{D}_j$ denote the representative publications of reviewer $r_j$, and let $x_p$ and $x_d$ denote the title-and-abstract text of submission $p$ and publication $d$. Reviewer-paper affinity is computed as 
    \begin{equation} A_{pj} = \max_{d\in\mathcal{D}_j} \  \cos\left( \operatorname{emb}(x_p), \operatorname{emb}(x_d) \right), 
    \label{eq:affinity} 
    \end{equation} 
    using the reviewer publication most aligned with the submission. We use $C_{pj}=1$ to denote a conflict of interest and exclude such pairs from bidding and assignment. The resulting substrate $(\mathcal{P},\mathcal{R},\mathbf{A},\mathbf{C},\Phi)$, rubric, and reference assessments are fixed across $z\in\{h,c\}$.

\subsection{Expertise-Grounded Collusion Construction}
\label{sec:collusion}
In the collusive world defined in \S\ref{sec:formulation}, a subset of reviewer-authors coordinates its bids to gain reciprocal access to one another's submissions. A meaningful instantiation of this threat model should preserve plausible reviewer-paper expertise relationships. Otherwise, colluders would target papers that they would be unlikely to bid on under honest behavior. \alg therefore derives both collusion rings and target papers from the fixed affinity and COI structure.

We consider reviewer-authors as candidate colluders because they possess submissions that can participate in reciprocal support. Let $\mathcal{P}_i$ denote the papers authored by reviewer $r_i$, and define the candidate set as
$V=\{r_i\in\mathcal{R}:\mathcal{P}_i\neq\emptyset\}$.
For each candidate, let
\[
    \mathcal{N}_K(i)
    :=
    \operatorname{TopK}_{p\in\mathcal{P}:C_{pi}=0} A_{pi}
\]
denote the reviewer's top-$K$ non-COI papers by affinity. Every paper in $\mathcal{N}_K(i)$ is therefore both eligible and compatible with the reviewer's expertise profile.

\textbf{Mutual-affinity graph.}
We construct an undirected graph $G=(V,E)$ over reviewer-authors, with
\begin{equation}
    (r_i,r_j)\in E
    \iff
    \big(
        \mathcal{P}_j\cap\mathcal{N}_K(i)\neq\emptyset
    \big)
    \land
    \big(
        \mathcal{P}_i\cap\mathcal{N}_K(j)\neq\emptyset
    \big).
    \label{eq:mutual_affinity}
\end{equation}
An edge requires reciprocal compatibility: each reviewer has at least one submission authored by the other in their own eligible high-affinity pool. This condition is stronger than one-way topical similarity and ensures that potential coordination is plausible in both directions. Importantly, $G$ is constructed only from authorship, $\mathbf{A}$, and $\mathbf{C}$, before any bids are generated; its edges therefore represent expertise compatibility rather than observed collusive behavior.

\textbf{Ring construction.}
Starting from $G$, we form small collusion rings through greedy clique expansion. A reviewer can be added to a ring only when they are connected to every existing member. Consequently, all reviewer pairs within a ring satisfy the mutual-affinity condition in Eq.~\eqref{eq:mutual_affinity}, providing pairwise expertise support for reciprocal bidding.

\textbf{Target construction.}
Targets are defined separately for each ring member. For reviewer $r_i$ in ring $\mathcal{R}_g$, we set
\begin{equation}
    \mathcal{T}_{g,i}
    =
    \Big(
        \bigcup\nolimits_{r_j\in\mathcal{R}_g\setminus\{r_i\}}
        \mathcal{P}_j
    \Big)
    \cap
    \mathcal{N}_K(i).
    \label{eq:ring_targets}
\end{equation}
Thus, $\mathcal{T}_{g,i}$ contains only papers authored by other ring members that already fall within $r_i$'s eligible high-affinity pool. The construction changes the reviewer's behavioral objective without fabricating reviewer-paper relevance. For each matched honest-collusive comparison, ring memberships and target sets are held fixed, but this coordination context is disclosed only to the collusive reviewer.

\subsection{Reviewer Simulacra}
\label{sec:simulacra}
    Given the expertise-grounded rings and target sets constructed in \S\ref{sec:collusion}, \alg instantiates each reviewer $r_j$ as an LLM-based reviewer simulacrum under behavioral world $z\in\{h,c\}$. The goal is not to reproduce a particular human reviewer, but to create an expertise-grounded role whose behavioral objective can be systematically controlled.
    
    \textbf{Layered reviewer personas.}
    Each persona separates fixed grounding from world-dependent behavioral control~\cite{li2026llm}. The grounding layer contains (i) the shared conference setting and reviewing rubric and (ii) the reviewer's research areas and representative publications. These components are identical for the same reviewer across behavioral worlds. The control layer specifies whether the reviewer behaves honestly or collusively, together with the corresponding bidding and reviewing objectives. In the collusive world, this layer additionally provides ring membership $\mathcal{R}_g$ and the reviewer-specific target set $\mathcal{T}_{g,i}$. Thus, the conference substrate and reviewer expertise remain unchanged, while only the information and objectives governing reviewer behavior vary.
    
    \textbf{Honest and collusive behavior.}
    Honest reviewer simulacra bid according to their expertise and interest and evaluate assigned papers independently under the shared rubric. They receive no information about collusion rings, target papers, or other reviewers' objectives. Collusive reviewer simulacra retain the same conference context, expertise profile, and rubric, but receive the coordination context defined above. During bidding, they strongly prioritize papers in $\mathcal{T}_{g,i}$ while continuing to bid by expertise on non-target papers. The resulting intervention therefore modifies preferences over reviewer-paper pairs that are already expertise-compatible, rather than introducing artificial relevance.
    
    \textbf{Strategic reviewing and stealth.}
    When a collusive reviewer is assigned a target paper, \alg controls the strength of downstream manipulation through the reviewing objective. An aggressive strategy seeks to strongly promote the target, whereas a subtle strategy seeks the most favorable evaluation that remains plausible under the shared rubric. Both strategies use the same target-seeking bidding policy; they differ only in how the reviewer evaluates a target after assignment. This separation allows \alg to distinguish whether coordinated bidding changes assignment access from how strategic reviewing subsequently affects evaluation.
    

\subsection{Bidding, Assignment, and Reviewing}
\label{sec:pathway}

\textbf{Bidding.}
    In world $z$, each reviewer $r_j$ submits an ordinal bid $B_{pj}^{(z)}$ for every $p\in\mathcal{N}_K(j)$ according to its honest or collusive objective. These bids form $\mathbf{B}^{(z)}$.

\textbf{Assignment.}
    After mapping ordinal bids to numerical utilities, the fixed assignment procedure scores each feasible pair and produces
    \begin{equation}
        u_{pj}^{(z)}
        =
        w_AA_{pj}+w_BB_{pj}^{(z)},
        \qquad
        \mathbf{M}^{(z)}
        =
        \Phi\big(
            \mathbf{B}^{(z)};\mathbf{A},\mathbf{C}
        \big),
        \quad
        M_{pj}^{(z)}\in\{0,1\}.
        \label{eq:assignment_utility}
    \end{equation}
    Here, $M_{pj}^{(z)}=1$ denotes an assignment. We use a lightweight greedy matcher that processes papers in a fixed order and assigns the highest-ranked eligible reviewers to each paper. Reviewers who have not yet satisfied their minimum service load receive priority until that obligation is met, after which assignments follow the standard affinity-bid ranking. 
    
    \textbf{Reviewing.}
    For each pair with $M_{pj}^{(z)}=1$, the reviewer $r_j$ produces a rubric-based review and recommendation score $Y_{pj}^{(z)}$, forming $\mathbf{Y}^{(z)}$. 
    \alg therefore preserves the sequential pathway $\mathbf{B}^{(z)}\rightarrow\mathbf{M}^{(z)}\rightarrow\mathbf{Y}^{(z)}$, allowing the experiments in \S\ref{sec:exp} to examine whether collusive bidding changes reviewer assignments and whether those assignment changes propagate into downstream evaluations.


\section{Experiments}
\label{sec:exp}

\subsection{Experimental Setup}
\label{sec:exp_setup}
\textbf{Conference instance and reference quality.}
We construct a fixed conference instance with $140$ researcher profiles from Semantic Scholar~\cite{ammar2018construction} and $100$ synthetic submissions spanning cs.LG, cs.CV, cs.CL, and cs.AI. Submissions are generated from author backgrounds under controlled quality conditions, with authors drawn from the reviewer pool, yielding $92$ distinct author-reviewers. An independent grading agent evaluates each paper from its title and abstract using the shared $1$-$10$ rubric, without access to preset quality, reviewer behavior, or collusion context. Its score $Q_p$ serves as the fixed reference-quality assessment and agrees substantially with the preset quality conditions (Pearson/Spearman $=0.832/0.828$). Reviewer-paper affinity is computed with \texttt{all-MiniLM-L6-v2}~\cite{reimers2019sentencebert} as the maximum cosine similarity over the reviewer's representative publications. Each reviewer bids on its top-$20$ non-COI papers, with paper authors excluded from bidding and assignment.

\textbf{Agents and assignment.}
All reviewer and grading agents use \texttt{deepseek-v4-flash}~\cite{deepseekai2026deepseekv4} with role-specific prompts. Bid labels from Very Low to Very High are encoded as $B_{pj}\in\{-100,-1,0,1,2\}$. A fixed greedy matcher assigns three reviewers per paper using
\[
    u_{pj}=A_{pj}+2B_{pj},
\]
with $B_{pj}=-100$ treated as a hard refusal. Reviewers have a maximum load of five papers, while author-reviewers must complete at least three reviews and receive priority until this minimum is met. Papers are processed by fixed reference-quality category, preserving their original order within each category. These categories are derived once from $Q_p$ and held fixed across behavioral worlds.

\textbf{Collusion and counterfactual conditions.}
We consider nominal collusion rates $r\in\{0,0.2,0.5\}$ over the $92$
author-reviewers, where $r=0$ denotes the all-honest condition. Following
\S\ref{sec:collusion}, colluders are organized into expertise-grounded rings of
two or three members, and each member targets only other members' papers within
its own top-$20$ affinity pool. The $r=0.2$ and $r=0.5$ conditions contain
$18.43\pm0.49$ and $46.71\pm0.70$ realized colluders, respectively. Rings use
either an aggressive or a subtle target-reviewing strategy. Each collusive
world is paired with an all-honest world that preserves the same papers,
reviewers, affinity scores, COI constraints, target relations, and assignment
mechanism; ring and target information is disclosed only to collusive agents.

\textbf{Runs and evaluation.}
We conduct seven runs over five structural seeds, with seed $0042$ independently
executed three times to assess LLM run-to-run variation under an identical
structural configuration. Papers, reviewer profiles, reference-quality
assessments, affinity scores, COI constraints, and matcher settings remain
fixed. Results report mean $\pm$ standard deviation over the seven runs. Additional
implementation details and agent prompts are provided in
Appx.~\ref{sec:appx-exp-setup}.

\textbf{Analysis structure.}
Our analysis follows the lifecycle in
Eq.~\eqref{eq:review_lifecycle}. We first examine whether coordinated
bids alter access to target papers
($\mathbf{B}^{(z)}\!\rightarrow\!\mathbf{M}^{(z)}$). We then measure whether
successful assignment capture changes target-paper evaluations
($\mathbf{M}^{(z)}\!\rightarrow\!\mathbf{Y}^{(z)}$). Finally, we assess whether
these localized effects produce a measurable conference-wide
footprint. \S\ref{sec:detectors} separately evaluates how much
evidence of the simulated attack is exposed to representative
bid-phase detectors.

\subsection{Effect on Assignment Access}
\label{sec:assignment_results}
Collusive reviewing is possible only if ring members obtain access to their
partners' submissions. For each paired run, let $\mathcal{S}$ contain every
directed \emph{support relation} $(r,p)$ for which reviewer $r$ is designated
to support a ring partner's target paper $p$. The same set $\mathcal{S}$ is
evaluated in the collusive world and its matched all-honest counterpart, thereby
isolating whether behavioral changes in bidding convert intended support into
assignments.

We measure access at two levels. The \emph{targeted assignment rate} (TAR)
uses support relations as its unit of analysis. The \emph{target-paper capture
rate} (TPCR) instead considers the set
$\mathcal{P}_T=\{p:\exists r,(r,p)\in\mathcal{S}\}$ and asks whether each target
paper is assigned at least one designated supporter:
\begin{equation}
\begin{aligned}
    \operatorname{TAR}^{(z)}
    &=
    \frac{1}{|\mathcal{S}|}
    \sum\nolimits_{(r,p)\in\mathcal{S}} M_{pr}^{(z)}, \\
    \operatorname{TPCR}^{(z)}
    &=
    \frac{1}{|\mathcal{P}_T|}
    \sum\nolimits_{p\in\mathcal{P}_T}
    \mathbb{I}\!\left[
        \sum\nolimits_{r:(r,p)\in\mathcal{S}}M_{pr}^{(z)}\geq1
    \right],
    \qquad z\in\{h,c\}.
\end{aligned}
\label{eq:assignment_access_metrics}
\end{equation}
where $\mathbb{I}[\cdot]$ is the indicator function, equal to one when
its argument is true and zero otherwise. TAR measures how often individual support relations are realized, whereas TPCR measures how often the ring gains any reviewing access to a target paper.

\noindent
\begin{minipage}[t]{0.53\linewidth}
    \vspace{0pt}
    Colluders place a Very High bid on every relation in $\mathcal{S}$, whereas
    only $16$-$18\%$ of the same relations receive a Very High bid under honest
    behavior. This intervention substantially, but not perfectly, converts into
    assignments. As shown in Fig.~\ref{fig:rq1_assignment}, TAR rises from
    $18.3$-$22.7\%$ under honest bidding to $62.4$-$64.1\%$ under collusive
    bidding, corresponding to gains of $41.4$ and $44.1$ percentage points at
    $r=0.2$ and $r=0.5$. TPCR similarly rises from $28.0$-$29.6\%$ to
    $71.8$-$76.9\%$, with gains of $42.2$ and $48.9$ points. Overall, coordinated bidding more than doubles assignment access. Approximately two-thirds of designated support relations are realized, and at least one colluder reaches roughly three-quarters of the targeted papers.

\end{minipage}
\hfill
\begin{minipage}[t]{0.44\linewidth}
    \vspace{0pt}
    \centering
    \includegraphics[width=\linewidth]
    {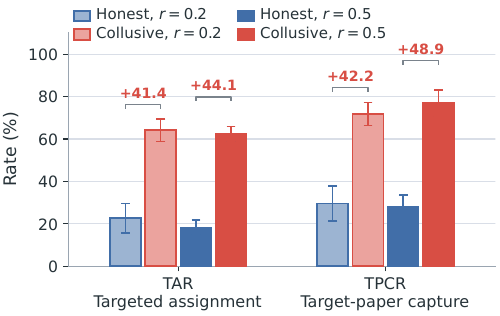}
    \refstepcounter{figure}
    \label{fig:rq1_assignment}
    {\footnotesize
    \textbf{Figure \thefigure.}
    Assignment access under honest and collusive bidding.
    Values are mean $\pm$ standard deviation over seven runs.}
\end{minipage}

\subsection{Target-Paper Score Inflation}
\label{sec:review_results}
The preceding results show that coordinated bids substantially increase
access to target papers. We next examine whether successful assignment
capture changes how these papers are evaluated. We distinguish
reviewer-level disagreement from the resulting paper-level score
change. At the reviewer level, we compare the score submitted by each assigned
supporting ring member with the mean score of the honest co-reviewers
assigned to the same captured paper. This within-paper comparison holds
the evaluated paper fixed and measures how differently the
strategically motivated reviewer evaluates it. The mean within-paper
differences are $2.21$ points at $r=0.2$ and $1.99$ points at
$r=0.5$.

Because every paper is assigned three reviewers, the denominator equals
three in our experiments. We define the matched paper-level score
change as
\begin{equation}
\Delta_p
=
\overline{Y}_p^{(c)}
-
\overline{Y}_p^{(h)},
\qquad
\overline{Y}_p^{(z)}
=
\frac{1}{3}
\sum\nolimits_{j\in\mathcal{R}}
M_{pj}^{(z)}Y_{pj}^{(z)}.
    \label{eq:paper_score_change}
\end{equation}
This quantity captures the combined downstream effect of altered reviewer
assignment and strategic reviewing. Consistent with the TPCR definition, a
target paper is classified as captured when at least one of its designated
supporting ring members is assigned in the collusive world.

\noindent
\begin{minipage}[t]{0.53\linewidth}
    \vspace{0pt}
    The reviewer-level discrepancy survives aggregation over the three reviews
    received by each paper. As shown in Fig.~\ref{fig:rq2_inflation}, target
    papers gain $0.61$-$0.69$ points on average relative to their matched
    all-honest outcomes. The change is concentrated among captured targets,
    whose means rise by $0.84$-$0.89$ points; uncaptured targets change by only
    $0.03$-$0.04$ points. Because capture is induced by the bidding
    intervention rather than independently randomized, this contrast provides
    mechanism-consistent localization rather than a separate causal estimate. 
    Reviewer- and strategy-specific analyses of target-paper score inflation are
    reported in Appx.~\ref{sec:appx-rq2-strategies}.
\end{minipage}
\hfill
\begin{minipage}[t]{0.44\linewidth}
    \vspace{0pt}
    \centering
    \includegraphics[width=\linewidth]
    {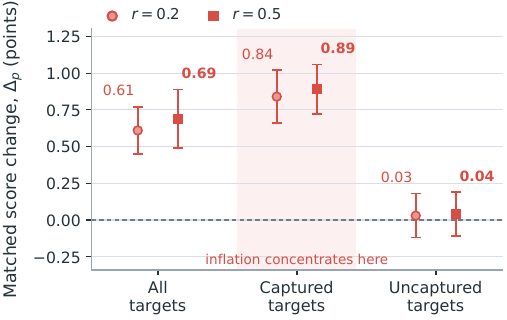}
    \refstepcounter{figure}
    \label{fig:rq2_inflation}
    {\footnotesize
    \textbf{Figure \thefigure.}
    Matched target-paper score changes across all, captured, and uncaptured
    targets.
    Values are mean $\pm$ standard deviation over seven runs.}
\end{minipage}

Taken together, these results show that the assignment effect propagates
into downstream evaluation. Once capture succeeds, supporting ring
members score the same papers approximately two points above honest
co-reviewers, increasing the captured paper's three-reviewer mean by
approximately $0.9$ points relative to its honest counterfactual.

\subsection{Conference-Wide Effects}
\label{sec:global_results}
The preceding analyses establish a localized pathway from coordinated
bids to assignment capture and target-paper score inflation. We now
assess whether these effects produce a measurable conference-wide
footprint along two complementary dimensions. \emph{Score inflation} measures changes in the distribution of the
$300$ individual review scores produced in each run.
\emph{Quality alignment} measures whether the three-reviewer paper
means remain aligned with the fixed reference-quality scores $Q_p$.
We quantify this alignment using Pearson correlation, Spearman rank
correlation, and overlap with the grader's top-$30$ papers. Let $s$
denote an individual review score. Reviews with $s>5.5$ are classified
as acceptance-level evaluations for this analysis; this threshold does
not represent the conference's final paper-acceptance decisions.

\begin{table}[t]
    \centering
    \small
    \caption{\textbf{Conference-wide score inflation and quality alignment.}
    Alignment metrics compare three-reviewer paper means with the fixed reference-quality
    assessments.}
    \label{tab:global_integrity}
    \begin{tabular}{@{}lccc@{}}
        \toprule
        Measurement & Honest & $r=0.2$ & $r=0.5$ \\
        \midrule

        \rowcolor{grayblue}
        \multicolumn{4}{l}{\textbf{Score inflation across all reviews}} \\

        Average review score
        & $4.34\pm0.02$
        & $4.47\pm0.06$
        & $4.72\pm0.09$ \\

        Reviews above threshold ($s>5.5$)
        & $28.2\pm0.4\%$
        & $30.6\pm1.7\%$
        & $34.2\pm2.1\%$ \\

        \addlinespace[3pt]

        \rowcolor{grayblue}
        \multicolumn{4}{l}{\textbf{Alignment with reference quality}} \\

        Pearson correlation
        & $0.804\pm0.010$
        & $0.775\pm0.019$
        & $0.775\pm0.025$ \\

        Spearman rank correlation
        & $0.813\pm0.014$
        & $0.776\pm0.021$
        & $0.765\pm0.035$ \\

        Overlap with grader top-$30$ ($/30$)
        & $23.0\pm0.0$
        & $22.3\pm0.7$
        & $22.9\pm0.3$ \\

        \bottomrule
    \end{tabular}
\end{table}

Collusion produces a clear, rate-dependent shift in conference-wide score
levels. Relative to the honest condition, the average review score increases by
$0.13$ points at $r=0.2$ and by $0.38$ points at $r=0.5$. The proportion of
reviews above the acceptance-level threshold likewise increases by $2.4$ and
$6.0$ percentage points. Thus, the localized target-paper effects identified in
\S\ref{sec:review_results} accumulate into broader score inflation as
the colluding population grows.

The effect on quality alignment is smaller. Pearson correlation decreases from
$0.804$ to $0.775$ under both collusion conditions, while Spearman correlation
decreases from $0.813$ to $0.776$ and $0.765$. In contrast, overlap with the
grader's top-$30$ remains between $22.3$ and $23.0$ papers across all
conditions. The results therefore indicate modest degradation in overall
score and rank alignment, but not a wholesale disruption of top-paper
membership.

In conclusion, target-level manipulation leaves a measurable but
limited conference-wide footprint. Score inflation increases with the
collusion rate, whereas reference-quality alignment weakens only
modestly and top-$30$ overlap remains nearly unchanged. The dominant
effect of collusion is therefore concentrated on successfully captured
targets, with a weaker aggregate effect at the conference level.

\section{Can \alg-generated Collusion be Detected?}
\label{sec:detectors}
We evaluate three bid-phase detector families on one fixed triplet comprising
an all-honest world and two collusive worlds. Collusion is restricted to the
92 author-reviewers in the 140-reviewer pool. The $r=0.2$ and $r=0.5$
instances contain 18 and 47 colluders arranged in 8 and 20 rings, respectively.
The detectors test reviewer anomaly rankings~\cite{jecmen2023dataset}, dense bid-author groups,
and dense reviewer-paper blocks~\cite{jecmen2025detection}. 
Tab.~\ref{tab:detector-summary} reports rank shifts for
rankings and recovered/flagged counts for sets. Assumptions, input mappings,
and full metrics appear in Appx.~\ref{sec:appx-detectors}.

\begin{table}[t]
    \centering
    \scriptsize
    \caption{\textbf{Representative bid-phase detector results.}
    $\Delta r$ is the collusive-minus-honest change in normalized rank
    (negative is more suspicious); set entries report recovered colluders /
    flagged reviewers. Complete results are provided in Appx.~\ref{sec:appx-detectors}.}
    \label{tab:detector-summary}
    \setlength{\tabcolsep}{3.5pt}
    \renewcommand{\arraystretch}{1.10}
    \begin{tabular}{@{}p{0.27\linewidth}p{0.25\linewidth}cc@{}}
        \toprule
        Input view & Detector
        & $r=0.2$ ($|\mathcal C|=18$)
        & $r=0.5$ ($|\mathcal C|=47$) \\
        \midrule

        \rowcolor{grayblue}
        \multicolumn{4}{l}{
            \textbf{Reviewer ranking}: $\Delta r$
        } \\
        Ternary bid matrix & Counting
        & $+0.054$ & $+0.025$ \\
        Ternary bid matrix & Pairwise
        & $-0.005$ & $-0.014$ \\
        Ternary bid matrix & Low-rank
        & $-0.032$ & $-0.024$ \\

        \addlinespace[2pt]
        \rowcolor{grayblue}
        \multicolumn{4}{l}{
            \textbf{Set recovery}: recovered / flagged reviewers
        } \\
        Bid-author, $\tau=1$ & Greedy densest
        & $16/89$ & $42/77$ \\
        Bid-author, $\tau=1$ & TellTail
        & $2.0/15.0$ & $12.3/16.7$ \\
        Bid-author, $\tau=2$ & TellTail
        & $3.2/4.2$ & $5.9/7.4$ \\
        Reviewer-paper, $\tau=1$ & Fraudar
        & $15/97$ & $41/101$ \\
        Reviewer-paper, $\tau=2$ & Fraudar
        & $6/36$ & $9/21$ \\

        \bottomrule
    \end{tabular}
\end{table}

The ranking methods provide little attack-induced separation. Counting's
rejection prior mismatches \alg's bid inflation, Pairwise Reciprocity largely
rediscovers reciprocity already induced by honest affinity, and Low-rank
Residual produces only small, encoding-dependent shifts.
On the native positive-bid view ($\tau=1$), OQC and TellTail change little from
the honest reference, while Densest and Fraudar obtain broad coverage only by
flagging $77$-$101$ of $140$ reviewers. At $\tau=2$, TellTail recovers one
complete ring with precision $.753/.874$, but recall remains $.178/.126$.
Each set detector returns one group against an attack distributed across eight
or 20 rings.

\textbf{Takeaway.}
\alg is not invisible: its strongest bids expose local structure. However, native positive-bid views are confounded by benign affinity, and single-set detectors do not cleanly recover the distributed collusive population. This conclusion concerns the evaluated detector-input combinations, not general undetectability.

\section{Conclusion}
\label{sec:conclusion}
We introduce \alg, a multi-agent simulacra framework for tracing expertise-grounded collusive bidding through reviewer assignment and downstream evaluation under matched behavioral worlds. Our experiments show that coordinated bids substantially increase colluders' access to target papers and, when assignment capture succeeds, lead to inflated evaluations of those papers, while aggregate conference-wide effects remain comparatively modest. Existing bid-phase detectors reveal a clear precision-coverage trade-off: native bid graphs often confound collusion with benign expertise-driven affinity, whereas stricter bid filtering improves localization but recovers only a small subset of colluders. Overall, \alg provides a controlled testbed for studying assignment-integrity risks and evaluating future defenses against collusive bidding.

\begin{ack}
This work was supported in part by Macau Science and Technology Development Fund under 001/2024/SKL, 0119/2024/RIB2 and 0110/2025/R1B2; in part by Research Committee at University of Macau under MYRG-CRG2025-00031-FST and MYRG-GRG2025-00086-FST; in part by the Guangdong Basic and Applied Basic Research Foundation under Grant 2024A1515012536.
\end{ack}


\bibliographystyle{unsrtnat}
\bibliography{ref}

@inproceedings{fiez2020super,
  title={A {SUPER*} algorithm to optimize paper bidding in peer review},
  author={Fiez, Tanner and Shah, Nihar and Ratliff, Lillian},
  booktitle={Proceedings of the 36th Conference on Uncertainty in Artificial Intelligence (UAI)},
  publisher={PMLR},
  pages={580--589},
  year={2020}
}

@inproceedings{wu2021making,
  title={Making paper reviewing robust to bid manipulation attacks},
  author={Wu, Ruihan and Guo, Chuan and Wu, Felix and Kidambi, Rahul and Van Der Maaten, Laurens and Weinberger, Kilian},
  booktitle={Proceedings of the 38th International Conference on Machine Learning (ICML)},
  publisher={PMLR},
  pages={11240--11250},
  year={2021}
}

@inproceedings{mimno2007expertise,
  title={Expertise modeling for matching papers with reviewers},
  author={Mimno, David and McCallum, Andrew},
  booktitle={Proceedings of the 13th ACM SIGKDD International Conference on Knowledge Discovery and Data Mining},
  pages={500--509},
  year={2007}
}

@article{charlin2012framework,
  title={A framework for optimizing paper matching},
  author={Charlin, Laurent and Zemel, Richard S and Boutilier, Craig},
  journal={arXiv preprint arXiv:1202.3706},
  year={2012}
}

@inproceedings{kobren2019paper,
  title={Paper matching with local fairness constraints},
  author={Kobren, Ari and Saha, Barna and McCallum, Andrew},
  booktitle={Proceedings of the 25th ACM SIGKDD International Conference on Knowledge Discovery \& Data Mining},
  pages={1247--1257},
  year={2019}
}

@inproceedings{rozencweig2023mitigating,
  title={Mitigating Skewed Bidding for Conference Paper Assignment},
  author={Rozencweig, Inbal and Meir, Reshef and Mattei, Nicholas and Amir, Ofra},
  booktitle={Proceedings of the 22nd International Conference on Autonomous Agents and Multiagent Systems (AAMAS)},
  pages={573--581},
  year={2023},
  publisher={IFAAMAS}
}

@inproceedings{lu2025agent,
  title={Agent reviewers: Domain-specific multimodal agents with shared memory for paper review},
  author={Lu, Kai and Xu, Shixiong and Li, Jinqiu and Ding, Kun and Meng, Gaofeng},
  booktitle={Proceedings of the 42nd International Conference on Machine Learning (ICML)},
  publisher={PMLR},
  year={2025}
}

@inproceedings{jecmen2023dataset,
  title={A dataset on malicious paper bidding in peer review},
  author={Jecmen, Steven and Yoon, Minji and Conitzer, Vincent and Shah, Nihar B and Fang, Fei},
  booktitle={Proceedings of the ACM Web Conference 2023},
  pages={3816--3826},
  year={2023}
}

@article{jecmen2025detection,
  title={On the Detection of Reviewer-Author Collusion Rings From Paper Bidding},
  author={Jecmen, Steven and Shah, Nihar B and Fang, Fei and Akoglu, Leman},
  journal={Transactions on Machine Learning Research},
  issn={2835-8856},
  year={2025}
}

@inproceedings{jecmen2020mitigating,
  title={Mitigating manipulation in peer review via randomized reviewer assignments},
  author={Jecmen, Steven and Zhang, Hanrui and Liu, Ryan and Shah, Nihar and Conitzer, Vincent and Fang, Fei},
  booktitle={Advances in Neural Information Processing Systems (NeurIPS)},
  volume={33},
  pages={12533--12545},
  year={2020}
}

@inproceedings{saveski2023counterfactual,
  title={Counterfactual evaluation of peer-review assignment policies},
  author={Saveski, Martin and Jecmen, Steven and Shah, Nihar and Ugander, Johan},
  booktitle={Advances in Neural Information Processing Systems (NeurIPS)},
  volume={36},
  pages={58765--58786},
  year={2023}
}

@inproceedings{boehmer2022combating,
  title={Combating collusion rings is hard but possible},
  author={Boehmer, Niclas and Bredereck, Robert and Nichterlein, Andr{\'e}},
  booktitle={Proceedings of the AAAI Conference on Artificial Intelligence},
  volume={36},
  number={5},
  pages={4843--4850},
  year={2022}
}

@inproceedings{hsieh2025vulnerability,
  title={Vulnerability of Text-matching in {ML/AI} conference reviewer assignments to collusions},
  author={Hsieh, Jhih-Yi Janet and Raghunathan, Aditi and Shah, Nihar B},
  booktitle={34th USENIX Security Symposium (USENIX Security 25)},
  pages={5189--5208},
  year={2025}
}

@inproceedings{jin2024agentreview,
  title={{AgentReview}: Exploring peer review dynamics with {LLM} agents},
  author={Jin, Yiqiao and Zhao, Qinlin and Wang, Yiyang and Chen, Hao and Zhu, Kaijie and Xiao, Yijia and Wang, Jindong},
  booktitle={Proceedings of the 2024 Conference on Empirical Methods in Natural Language Processing (EMNLP)},
  publisher={Association for Computational Linguistics},
  pages={1208--1226},
  year={2024}
}

@inproceedings{goldsmith2007ai,
  title={The {AI} conference paper assignment problem},
  author={Goldsmith, Judy and Sloan, Robert H},
  booktitle={Proceedings of the AAAI Workshop on Preference Handling for Artificial Intelligence},
  pages={53--57},
  year={2007}
}

@article{stelmakh2021peerreview4all,
  title={PeerReview4All: Fair and accurate reviewer assignment in peer review},
  author={Stelmakh, Ivan and Shah, Nihar and Singh, Aarti},
  journal={Journal of Machine Learning Research},
  volume={22},
  number={163},
  pages={1--66},
  year={2021}
}

@inproceedings{xu2023one,
  title={A one-size-fits-all approach to improving randomness in paper assignment},
  author={Xu, Yixuan and Jecmen, Steven and Song, Zimeng and Fang, Fei},
  booktitle={Advances in Neural Information Processing Systems (NeurIPS)},
  volume={36},
  pages={14445--14468},
  year={2023}
}

@article{liang2024monitoring,
  title={Monitoring {AI}-modified content at scale: A case study on the impact of {ChatGPT} on {AI} conference peer reviews},
  author={Liang, Weixin and Izzo, Zachary and Zhang, Yaohui and Lepp, Haley and Cao, Hancheng and Zhao, Xuandong and Chen, Lingjiao and Ye, Haotian and Liu, Sheng and Huang, Zhi and others},
  journal={arXiv preprint arXiv:2403.07183},
  year={2024}
}

@inproceedings{zhou2024llm,
  title={Is {LLM} a reliable reviewer? A comprehensive evaluation of {LLM} on automatic paper reviewing tasks},
  author={Zhou, Ruiyang and Chen, Lu and Yu, Kai},
  booktitle={Proceedings of the 2024 Joint International Conference on Computational Linguistics, Language Resources and Evaluation (LREC-COLING 2024)},
  pages={9340--9351},
  year={2024}
}

@inproceedings{yu2024automated,
  title={Automated peer reviewing in paper sea: Standardization, evaluation, and analysis},
  author={Yu, Jianxiang and Ding, Zichen and Tan, Jiaqi and Luo, Kangyang and Weng, Zhenmin and Gong, Chenghua and Zeng, Long and Cui, Renjing and Han, Chengcheng and Sun, Qiushi and others},
  booktitle={Findings of the Association for Computational Linguistics: EMNLP 2024},
  publisher={Association for Computational Linguistics},
  pages={10164--10184},
  year={2024}
}

@inproceedings{park2023generative,
  title={Generative agents: Interactive simulacra of human behavior},
  author={Park, Joon Sung and O'Brien, Joseph and Cai, Carrie Jun and Morris, Meredith Ringel and Liang, Percy and Bernstein, Michael S},
  booktitle={Proceedings of the 36th Annual ACM Symposium on User Interface Software and Technology (UIST)},
  pages={1--22},
  year={2023}
}

@inproceedings{schmidgall2025agent,
  title={Agent laboratory: Using {LLM} agents as research assistants},
  author={Schmidgall, Samuel and Su, Yusheng and Wang, Ze and Sun, Ximeng and Wu, Jialian and Yu, Xiaodong and Liu, Jiang and Moor, Michael and Liu, Zicheng and Barsoum, Emad},
  booktitle={Findings of the Association for Computational Linguistics: EMNLP 2025},
  publisher={Association for Computational Linguistics},
  pages={5977--6043},
  year={2025}, 
}

@misc{aaai2026xreports,
  author = {{Association for the Advancement of Artificial Intelligence}},
  title = {{AAAI-27 Statement on Potential Collusion During Reviewer Bidding}},
  year = {2026},
  month = jul,
  howpublished = {Post on X, \url{https://x.com/RealAAAI/status/2082108476302479560}},
  note = {Posted July 28, 2026; accessed September 4, 2026}
}

@misc{aaai2026biddingweb,
  author = {{Association for the Advancement of Artificial Intelligence}},
  title = {{AAAI-27 Letter: Reviewer Bidding Integrity}},
  year = {2026},
  month = aug,
  howpublished = {AAAI Publication Policies and Guidelines,
  \url{https://aaai.org/aaai-publications/aaai-publication-policies-guidelines/}},
  note = {Accessed September 4, 2026}
}

@inproceedings{reimers2019sentencebert,
  title={Sentence-{BERT}: Sentence Embeddings using Siamese {BERT}-Networks},
  author={Reimers, Nils and Gurevych, Iryna},
  booktitle={Proceedings of the 2019 Conference on Empirical Methods in Natural Language Processing and the 9th International Joint Conference on Natural Language Processing (EMNLP-IJCNLP)},
  pages={3982--3992},
  year={2019},
  publisher={Association for Computational Linguistics}
}

@misc{deepseekai2026deepseekv4,
  title={{DeepSeek-V4}: Towards Highly Efficient Million-Token Context Intelligence},
  author={{DeepSeek-AI}},
  year={2026},
  note={Technical report}
}

@inproceedings{anjum2019pare,
  title={{PaRe}: A Paper-Reviewer Matching Approach Using a Common Topic Space},
  author    = {Anjum, Omer and Gong, Hongyu and Bhat, Suma and Hwu, Wen-Mei and Xiong, JinJun},
  booktitle = {Proceedings of the 2019 Conference on Empirical Methods in Natural Language Processing and the 9th International Joint Conference on Natural Language Processing (EMNLP-IJCNLP)},
  pages     = {518--528},
  year      = {2019},
  publisher = {Association for Computational Linguistics},
}

@inproceedings{ammar2018construction,
  title= {Construction of the Literature Graph in {Semantic Scholar}},
  author    = {Ammar, Waleed and Groeneveld, Dirk and Bhagavatula, Chandra and Beltagy, Iz and Crawford, Miles and Downey, Doug and Dunkelberger, Jason and Elgohary, Ahmed and Feldman, Sergey and Ha, Vu and Kinney, Rodney and Kohlmeier, Sebastian and Lo, Kyle and Murray, Tyler and Ooi, Hsu-Han and Peters, Matthew and Power, Joanna and Skjonsberg, Sam and Wang, Lucy and Wilhelm, Chris and Yuan, Zheng and van Zuylen, Madeleine and Etzioni, Oren},
  booktitle = {Proceedings of the 2018 Conference of the North American Chapter of the Association for Computational Linguistics: Human Language Technologies, Volume 3 (Industry Papers)},
  pages     = {84--91},
  year      = {2018},
  publisher = {Association for Computational Linguistics},
}

@article{li2026llm,
  title={{LLM} Persona Unlearning},
  author={Li, Kemou and Shi, Zhuan and Wang, Qizhou and Li, Fengpeng and Rostamzadeh, Negar and Farnadi, Golnoosh and Zhou, Jiantao},
  journal={arXiv preprint arXiv:2609.39882},
  year={2026}
}

\appendix

\begin{center}
    \LARGE \bf Appendix of \alg
\end{center}

\definecolor{kleinblue}{rgb}{0,0.18,0.65}

\etocdepthtag.toc{mtappendix}
\etocsettagdepth{mtchapter}{none}
\etocsettagdepth{mtappendix}{subsection}
{
  \hypersetup{linkcolor=kleinblue}
  \tableofcontents
}

\clearpage
\section*{Overview of the Appendix}

The appendix is organized as follows:
\begin{itemize}[leftmargin=*,itemsep=0.12em,topsep=0.2em]

    \item \S\ref{sec:appx-related-work} reviews prior work on reviewer
    assignment, strategic manipulation, collusion detection, and LLM-based
    peer-review agents.

    \item \S\ref{sec:appx-exp-setup} provides further experimental details,
    including conference and submission construction, reference-quality
    assessment, agent prompting, assignment implementation, collusion
    configurations, and repeated runs.

    \item \S\ref{sec:appx-additional-results} documents the provenance,
    cross-artifact structure, integrity checks, and responsible-release protocol
    for the experimental data, followed by supplementary analyses of assignment
    access, target-paper score inflation, and conference-wide effects.

    \item \S\ref{sec:appx-detectors} specifies the detector families, input
    representations, evaluation protocol, reporting measures, and complete
    detector-level results.

    \item \S\ref{sec:appx-discussion} discusses the implications of the
    findings for collusion detection and reviewer-assignment mechanism design.

    \item \S\ref{sec:appx-limitations} summarizes the scope of the current
    simulation and evaluation and outlines directions for future work.

    \item \S\ref{sec:appx-broader-impacts} discusses the potential benefits of
    peer-review integrity research together with its dual-use, false-positive,
    fairness, and privacy risks.

\end{itemize}

\section{Related Work}
\label{sec:appx-related-work}
\subsection{Reviewer Assignment and Bidding Mechanisms}
    Reviewer-paper matching is a fundamental component of modern peer-review
    systems \cite{stelmakh2021peerreview4all}. Early work focused on estimating reviewer
    expertise and paper-reviewer affinity from publication histories, topic models,
    and textual similarity \cite{mimno2007expertise,charlin2012framework}.
    Anjum et al.~\cite{anjum2019pare} proposed \textit{PaRe}, which matches papers
    and reviewer profiles in a shared topic space to mitigate vocabulary mismatch
    and partial topic overlap. More recent approaches formulate reviewer assignment
    as a constrained optimization problem, balancing matching quality with reviewer
    workload, fairness, and allocation constraints
    \cite{goldsmith2007ai,kobren2019paper,stelmakh2021peerreview4all}.
    
    Beyond automatically inferred affinity scores, reviewer-provided preferences have become an important source of information in practical conference assignment pipelines. In particular, bidding mechanisms allow reviewers to explicitly indicate their interest and willingness to review specific submissions, providing complementary signals for optimization-based assignment. 
    Fiez et al.~\cite{fiez2020super} optimized paper
    presentation during bidding to improve bid coverage, while Rozencweig et
    al.~\cite{rozencweig2023mitigating} investigated interventions for reducing
    skewed bidding and orphan papers. Saveski et
    al.~\cite{saveski2023counterfactual} further developed counterfactual methods
    for evaluating alternative reviewer-assignment policies from randomized
    assignments. Recent work has also investigated scalable assignment policies for
    large conference settings \cite{xu2023one}. Overall, these methods primarily
    treat reviewer expertise, preferences, and bids as signals for constructing or
    evaluating assignments, rather than as strategically generated behaviors whose
    effects propagate through subsequent review stages.
    
\subsection{Strategic Manipulation and Adversarial Behaviors in Peer Review}

    A growing body of work has examined the vulnerability of peer-review mechanisms
    to strategic and adversarial reviewer behavior. Jecmen et
    al.~\cite{jecmen2020mitigating} studied malicious reviewer targeting, including
    quid-pro-quo and torpedo-reviewing scenarios, and proposed randomized assignment
    mechanisms that limit the probability of obtaining strategically desired
    assignments. Wu et al.~\cite{wu2021making} focused directly on bid manipulation,
    showing how strategically altered reviewer preferences can influence assignment
    and developing mechanisms to improve robustness against such attacks.
    
    Subsequent work has examined both empirical attack behavior and structural forms
    of collusion. Jecmen et al.~\cite{jecmen2023dataset} released a dataset of
    malicious paper bidding collected through a controlled mock-conference activity
    and analyzed how different bidding strategies affect reviewer assignment.
    Boehmer et al.~\cite{boehmer2022combating} studied cycle-free reviewer
    assignments designed to prevent reciprocal reviewer-author structures, while
    Jecmen et al.~\cite{jecmen2025detection} investigated the detection of
    reviewer-author collusion rings from bidding signals. Hsieh et
    al.~\cite{hsieh2025vulnerability} further showed that collusive assignment
    manipulation can arise through reviewer-paper text-matching signals even in the
    absence of bidding.
    
    These studies establish that both bids and matching signals can create
    meaningful assignment-integrity risks. However, they predominantly characterize
    attack behaviors, detect suspicious structures, or redesign the assignment
    stage itself. They do not jointly model how expertise-grounded collusive
    behavior generates strategic bids, changes reviewer assignments, and
    subsequently alters the reviews produced for the same target papers under a
    matched conference environment.

\subsection{LLM-based Peer Review Agents and Simulation}
    
    Recent advances in large language models have enabled both automated review
    generation and agent-based simulation of academic peer review. Initial studies
    explored LLMs for generating review reports, assessing scientific manuscripts,
    and supporting review-quality evaluation
    \cite{liang2024monitoring,zhou2024llm,yu2024automated}.
    
    More recent work has moved from single-model review generation toward explicit
    reviewer agents. Jin et al.~\cite{jin2024agentreview} introduced
    \textit{AgentReview}, an LLM-based peer-review simulation framework that models
    reviewer and decision-making roles and enables controlled analysis of latent
    behavioral factors and their effects on reviews and paper decisions. Lu et
    al.~\cite{lu2025agent} developed domain-specific multimodal reviewer agents with
    shared memory to provide more specialized and context-aware paper reviews.
    
    These approaches demonstrate that LLM agents can reproduce important aspects of
    reviewer evaluation and support controlled studies of peer-review dynamics.
    However, existing agent-based peer-review systems largely begin after reviewer
    access to papers has already been determined: they do not jointly model
    strategic bidding, endogenous reviewer assignment, and the resulting downstream
    reviews. Beyond peer review, broader studies have demonstrated the potential of LLM agents as programmable simulations of human behavior and social interactions \cite{park2023generative,schmidgall2025agent}.

\section{Further Experimental Setup}
\label{sec:appx-exp-setup}

\subsection{Conference and Submission Construction}
\label{sec:appx-conference-construction}

The reviewer pool contains $140$ Semantic Scholar researcher profiles from four
computer-science areas: cs.LG ($37$), cs.CV ($36$), cs.CL ($34$), and cs.AI
($33$). Each profile contains the researcher's areas and representative
publications, which are used to ground the corresponding reviewer agent.

The submission set contains $100$ synthetic papers generated from author
backgrounds under four preset quality conditions: $5$ strong accept, $25$
accept, $40$ borderline, and $30$ reject. It comprises $60$ single-author
papers, $30$ dual-author papers whose authors belong to the same area, and $10$
dual-author papers with cross-area author pairs. Authors are sampled from the
reviewer pool, producing $92$ distinct author-reviewers. These papers, authors,
and reviewer profiles are fixed across all behavioral worlds and runs.

For each paper-reviewer pair, affinity is computed with \texttt{all-MiniLM-L6-v2}~\cite{reimers2019sentencebert}. We separately embed the title-and-abstract text of the submission and each representative publication in the reviewer profile, and define affinity as the maximum cosine similarity over that reviewer's publications, as specified in Eq.~\eqref{eq:affinity}. Each reviewer retains its top-$20$ non-COI papers as its bidding pool.

\subsection{Reference-Quality Assessment}
\label{sec:appx-reference-quality}

The quality-grading agent uses the same LLM backbone as the reviewer agents but
receives an independent system prompt describing the role of a calibrated
senior area chair. It evaluates papers comparatively from their titles and
abstracts using the shared $1$-$10$ rubric and is instructed to calibrate its
scores against a conference acceptance rate of approximately $30\%$. The grader
does not receive the papers' preset quality conditions or any reviewer,
bidding, assignment, or collusion information.

Let $Q_p\in[1,10]$ denote the grader's continuous assessment of paper $p$.
These scores are generated freely by the LLM and are not constrained to follow
a predetermined distribution. For analyses requiring categorical reference
labels, we sort papers by $Q_p$ and apply fixed quantile thresholds:
the top $5\%$ are labeled strong accept, the next $25\%$ accept, the next
$40\%$ borderline, and the remaining $30\%$ reject. Thus, the categorical
labels follow the $5/25/40/30$ distribution by construction, whereas the
continuous scores remain unconstrained.

Although the grader does not observe the preset generation conditions, its
continuous assessments show substantial agreement with them
(Pearson/Spearman $=0.832/0.828$). Both the continuous scores and the derived
labels are computed once and then fixed across all experimental runs.

\subsection{Agent Prompting and Generation Configuration}
\label{sec:appx-agent-prompts}

Each agent prompt is assembled from a shared scaffold and a
behavior-specific policy. The shared reviewer scaffold contains the conference
context, the common $1$-$10$ scoring rubric, the reviewer's research areas,
and three representative publication titles. The quality grader receives the
same conference context and rubric, but an independent role description.
Tab.~\ref{tab:agent-prompt-excerpts} summarizes the prompt structure and
shows the instructions that define the experimental behaviors.

\begin{table}[t]
    \centering
    \footnotesize
    \caption{\textbf{Prompt scaffold and selected behavior-defining excerpts.}
    Runtime fields such as research areas, ring peers, and target-paper IDs are
    instantiated separately for each agent. Repeated paper formatting and JSON
    output instructions are omitted for clarity.}
    \label{tab:agent-prompt-excerpts}
    \setlength{\tabcolsep}{5pt}
    \renewcommand{\arraystretch}{1.16}
    \begin{tabular}{@{}
        >{\raggedright\arraybackslash}p{0.20\linewidth}
        >{\raggedright\arraybackslash}p{0.74\linewidth}
        @{}}
        \toprule
        Prompt component & Content supplied to the agent \\
        \midrule

        \rowcolor{grayblue}
        \multicolumn{2}{l}{\textbf{Shared prompt scaffold}} \\

        Reviewer context
        & Conference context and score calibration; the shared scoring rubric;
        the reviewer's research areas; and three representative publication
        titles. \\

        Quality grader
        & \emph{``You assess the quality of all submitted papers together,
        comparatively, based on their title and abstract. You must rank them
        and assign each a score that reflects the conference's rigorous
        standards and acceptance rate.''} \\

        \addlinespace[2pt]
        \rowcolor{grayblue}
        \multicolumn{2}{l}{\textbf{Behavior-specific policy}} \\

        Honest reviewer
        & \emph{``Evaluate papers based solely on their technical quality,
        originality, clarity, and significance \ldots{} calibrate your scores
        honestly.''} \\

        Collusive bidding
        & After receiving its ring peers and target-paper IDs, the agent is
        instructed: \emph{``You MUST bid `Very High' on every paper in your
        ring's target list. For all other papers, bid according to your genuine
        research expertise.''} \\

        Target reviewing
        & \textbf{Aggressive:} \emph{``Give it a score STRICTLY ABOVE 6
        \ldots{} chosen as the highest score you can still defend from the
        paper's content.''}
        \newline
        \textbf{Subtle:} \emph{``Give it the HIGHEST score an honest, fair
        expert in your area could plausibly give \ldots{} Keep your review
        plausible and do not mention the ring.''} \\

        \bottomrule
    \end{tabular}
\end{table}

The shared scaffold is held fixed across honest and collusive worlds. Collusive
agents additionally receive their ring membership and reviewer-specific target
papers. Both aggressive and subtle agents receive the same target-bidding
instruction; the two policies differ only in how they evaluate a target after
being assigned to it. Subtle agents are further instructed to vary their
language across ring papers so that coordination is not explicit in the review
text.

For bidding, papers are presented in chunks of ten and the agent must select
one of five labels from Very Low to Very High. Assigned papers are reviewed in
a batch using the shared rubric, with a numerical score and concise review
requested for each paper. Bids and reviews are returned as structured JSON.

Reviewer bidding and reviewing use temperature $0.4$, while reference-quality
grading uses temperature $0.2$; all calls use
\texttt{deepseek-v4-flash} with a maximum of $4096$ output tokens. Invalid or
incomplete outputs are retried. After the retry limit, missing bids are filled
with Neutral and missing reviews with score $5$; a failed batch review is first
retried as separate single-paper calls.




\subsection{Bidding and Assignment Implementation}
\label{sec:appx-assignment-details}

Each reviewer bids on its top-$20$ non-COI papers using one of five ordinal
labels. The matcher encodes these labels as
\[
\begin{array}{c|ccccc}
\text{Label}
& \text{Very Low}
& \text{Low}
& \text{Neutral}
& \text{High}
& \text{Very High} \\
\hline
B_{pj}
& -100
& -1
& 0
& 1
& 2
\end{array}
\]
and computes the base pairwise utility
\[
    u_{pj}=A_{pj}+2B_{pj}.
\]
The Very Low value produces a strongly negative utility and is treated as a
hard refusal. The greedy matcher processes papers according to the fixed reference-quality category order: strong accept, accept, borderline, and reject. These categories are obtained by ranking the continuous reference-quality scores ($Q_p$) and applying the fixed (5/25/40/30) quantile partition. Within each category, the matcher preserves the original input order. The resulting order is fixed across behavioral worlds and affects only when papers access scarce reviewer capacity; neither the category labels nor ($Q_p$) otherwise enter the assignment objective.

Each paper receives three eligible reviewers. Every reviewer has a maximum load
of five papers, and author-reviewers have a minimum service load of three.
During matching, eligible author-reviewers below this minimum receive dominant
priority until the obligation is satisfied; subsequent assignments follow the
affinity-bid utility ranking.

\subsection{Collusion Configurations and Repeated Runs}
\label{sec:appx-collusion-runs}

Collusion is restricted to the $N_a=92$ author-reviewers. For nominal rate $r$,
the target number of colluders is
\[
    n_c=\left\lfloor rN_a\right\rfloor.
\]
Complete rings are retained during construction, so the realized number may
differ slightly from this quota. Across the seven runs, the $r=0.2$ and
$r=0.5$ conditions contain $18.43\pm0.49$ and $46.71\pm0.70$ colluders,
respectively.

Rings are greedily constructed as cliques in the mutual top-$20$ affinity graph
defined in \S\ref{sec:collusion}. The implementation produces rings containing
two or three members. Each member's targets are restricted to papers authored
by other ring members that also occur in its own non-COI top-$20$ bidding pool.

Each ring, rather than each individual reviewer, is independently assigned an
aggressive or subtle target-reviewing strategy with equal probability. The
realized strategy proportions therefore vary across runs. Tab.~\ref{tab:appx-strategy-allocation} reports the resulting numbers of aggressive
and subtle colluders.

\begin{table}[t]
    \centering
    \small
    \caption{\textbf{Realized ring-level reviewing strategies.}
    Each entry reports the number of aggressive/subtle colluders. All members
    of the same ring share its assigned strategy.}
    \label{tab:appx-strategy-allocation}
    \setlength{\tabcolsep}{7pt}
    \begin{tabular}{lcc}
        \toprule
        Run & $r=0.2$ & $r=0.5$ \\
        \midrule
        $0042$-1 & $14/4$  & $22/25$ \\
        $0042$-2 & $13/6$  & $24/23$ \\
        $0042$-3 & $14/4$  & $18/28$ \\
        $0921$    & $12/7$  & $34/13$ \\
        $1125$    & $11/7$  & $25/23$ \\
        $2024$    & $7/11$  & $21/25$ \\
        $7777$    & $9/10$  & $20/26$ \\
        \bottomrule
    \end{tabular}
\end{table}

We use five structural seeds, $\{0042,0921,1125,2024,7777\}$. Seed $0042$ is
executed three times with the same structural configuration but independent LLM
calls, yielding seven runs in total. The remaining seeds vary ring construction
and ring-level reviewing-strategy assignment. Papers, reviewer profiles,
reference-quality assessments, affinities, COI constraints, and matcher
hyperparameters remain fixed. Reported means and standard deviations are
computed over all seven runs.

\section{Additional Experimental Details}
\label{sec:appx-additional-results}

\subsection{Data Provenance, Structure, and Release}
\label{sec:appx-data-artifacts}
The conference substrate is grounded in researcher profiles collected from
public Semantic Scholar metadata. The resulting snapshot contains $140$
pseudonymized researcher profiles across cs.LG ($37$), cs.CV ($36$), cs.CL
($34$), and cs.AI ($33$). Each profile records research areas and representative
publications and is used to ground the expertise of one reviewer simulacrum.
Among these profiles, $92$ are selected as reviewer-authors and provide the
research context from which the $100$ synthetic submissions are generated.
Thus, the reviewer and author profiles originate from real public metadata,
whereas the submissions, bids, assignments, and reviews are produced within
the simulated conference.

Fig.~\ref{fig:data-provenance} shows how the collected profiles are transformed
into the fixed conference substrate and world-specific behavioral artifacts.
The raw profile snapshot is treated as restricted data because combinations of
publication histories and research areas may permit re-identification, even
when direct names are removed.

\begin{figure*}[t]
    \centering

    \begin{tikzpicture}[
        artifact/.style={
            draw=black!55,
            rounded corners=2pt,
            align=center,
            font=\footnotesize,
            text width=0.205\textwidth,
            minimum height=1.15cm,
            inner sep=4pt
        },
        public/.style={
            artifact,
            fill=green!7
        },
        restricted/.style={
            artifact,
            fill=orange!12
        },
        derived/.style={
            artifact,
            fill=grayblue
        },
        flow/.style={
            ->,
            thick,
            draw=black!65,
            >=stealth
        }
    ]

        \node[public] (source)
        at (0.125\textwidth,1.7) {
            \textbf{Public metadata}\\
            Semantic Scholar\\
            researcher records
        };

        \node[restricted] (profiles)
        at (0.375\textwidth,1.7) {
            \textbf{Frozen profiles}\\
            140 real researchers \\
            profiles
        };

        \node[derived] (papers)
        at (0.625\textwidth,1.7) {
            \textbf{Synthetic papers}\\
            100 submissions\\
            92 reviewer-authors
        };

        \node[derived] (substrate)
        at (0.875\textwidth,1.7) {
            \textbf{Fixed substrate}\\
            14,000 affinities\\
            authorship and COIs
        };

        \node[derived] (personas)
        at (0.125\textwidth,-0.2) {
            \textbf{Reviewer personas}\\
            140 per world\\
            honest or collusive
        };

        \node[derived] (bids)
        at (0.375\textwidth,-0.2) {
            \textbf{Bids}\\
            2,800 per world\\
            five ordinal levels
        };

        \node[derived] (assignments)
        at (0.625\textwidth,-0.2) {
            \textbf{Assignments}\\
            300 per world\\
            three per paper
        };

        \node[derived] (reviews)
        at (0.875\textwidth,-0.2) {
            \textbf{Reviews and metrics}\\
            300 reviews per world\\
            multi-level outcomes
        };

        \draw[flow] (source) -- (profiles);
        \draw[flow] (profiles) -- (papers);
        \draw[flow] (papers) -- (substrate);

        \draw[flow] (personas) -- (bids);
        \draw[flow] (bids) -- (assignments);
        \draw[flow] (assignments) -- (reviews);

        \draw[flow] (profiles.south) -- (personas.north);
        \draw[flow] (papers.south) -- (bids.north);
        \draw[flow] (substrate.south) -- (assignments.north);

    \end{tikzpicture}

    \caption{
        \textbf{Data provenance and artifact organization in \alg.}
        Public researcher metadata is collected into a frozen profile snapshot
        that grounds both reviewer expertise and synthetic submission generation.
        The conference substrate is held fixed across behavioral worlds, while
        personas, bids, assignments, reviews, and evaluation metrics are generated
        separately for each world. Orange denotes the restricted profile snapshot;
        blue denotes synthetic or derived artifacts that can be released after
        consistent identifier remapping and privacy review.
    }
    \label{fig:data-provenance}
\end{figure*}

Figs.~\ref{fig:data-schema-substrate} and
\ref{fig:data-schema-worlds} show schema-faithful excerpts from the corresponding
JSON artifacts. Field names and nesting match the files used in our experiments.
Values are shortened or replaced by descriptive placeholders for readability
and privacy. In particular, publication titles, abstracts, institutions, and
coauthor relations from collected profiles are not reproduced.

\begin{figure*}[!t]
\centering

\begin{minipage}[t]{0.485\textwidth}
\textbf{Collected profile: \texttt{profiles\_140.json}}
\begin{lstlisting}[style=artifactjson]
[
  {
    "profile_id": "U_<random_id>",
    "research_areas": ["cs.CV", "vision", "..."],
    "sub_areas": ["segmentation", "..."],
    "publications": [
      {
        "title": "[withheld]",
        "abstract": "[withheld]",
        "venue": "[withheld]",
        "year": 2023,
        "keywords": ["vision", "..."],
        "venue_type": "conference",
        "citation_count": 120
      }
    ],
    "h_index_estimate": 12,
    "total_papers_estimate": 10,
    "institutions": ["[withheld]"],
    "coauthor_ids": ["U_<random_id>"],
    "source_type": "semantic_scholar",
    "source_timestamp": "2026-08-17T..."
  }
]
\end{lstlisting}
\end{minipage}
\hfill
\begin{minipage}[t]{0.485\textwidth}
\textbf{Synthetic submission: \texttt{papers\_graded.json}}
\begin{lstlisting}[style=artifactjson]
[
  {
    "paper_id": "P0001",
    "title": "[synthetic title]",
    "abstract": "[synthetic abstract]",
    "keywords": [
      "video editing",
      "diffusion models",
      "reinforcement learning"
    ],
    "primary_area": "cs.CV",
    "preset_quality": "borderline",
    "author_type": "single",
    "author_profile_ids": ["U_<random_id>"],
    "generated_at": "2026-08-19T...",
    "generation_model": "llm",
    "quality": "borderline",
    "grader_score": 4.0,
    "grader_reasoning": ""
  }
]
\end{lstlisting}
\end{minipage}

\vspace{0.65em}

\begin{minipage}[t]{0.485\textwidth}
\textbf{Affinity matrix: \texttt{affinity\_100.json}}
\begin{lstlisting}[style=artifactjson]
{
  "model": "all-MiniLM-L6-v2",
  "num_papers": 100,
  "num_reviewers": 140,
  "num_scores": 14000,
  "affinity_stats": {
    "mean": 0.3341,
    "min": -1.0,
    "max": 0.7331
  },
  "scores": [
    {
      "paper_id": "P0065",
      "reviewer_id": "U_<random_id>",
      "score": 0.7331
    }
  ]
}
\end{lstlisting}
\end{minipage}
\hfill
\begin{minipage}[t]{0.485\textwidth}
\textbf{Derived metrics: \texttt{paper\_level\_metrics.json}}
\begin{lstlisting}[style=artifactjson]
{
  "target_score_deviation": 1.4094,
  "deviation_by_quality": {
    "strong_accept": 0.1944,
    "accept": -0.0870,
    "borderline": 1.6671,
    "reject": 0.8580
  },
  "n_target_papers": 26,
  "n_nontarget_papers": 74,
  "per_paper": {
    "P0001": {
      "avg_score": 2.83,
      "is_target": true,
      "quality": "borderline"
    }
  }
}
\end{lstlisting}
\end{minipage}

\caption{Representative JSON structures for the fixed conference substrate and
derived paper-level outputs. The collected researcher profiles are based on
real public metadata, whereas submission contents are synthetically generated.
Profile values shown here are masked; field names and data types follow the
experimental artifacts.}
\label{fig:data-schema-substrate}
\end{figure*}

\begin{figure*}[!t]
\centering

\begin{minipage}[t]{0.485\textwidth}
\textbf{Reviewer persona: \texttt{personas.json}}
\begin{lstlisting}[style=artifactjson]
[
  {
    "reviewer_id": "U_<random_id>",
    "behavior_type": "collusion",
    "research_areas": ["cs.CL", "..."],
    "sample_publications": ["[withheld]", "..."],
    "stealth": "aggressive",
    "ring_id": "ring_A",
    "ring_peers": [
      "U_<random_id>",
      "U_<random_id>"
    ],
    "target_paper_ids": [
      "P0034", "P0042", "P0078"
    ],
    "target_description":
      "papers authored by ring members"
  }
]
\end{lstlisting}
\end{minipage}
\hfill
\begin{minipage}[t]{0.485\textwidth}
\textbf{Bid record: \texttt{bids.json}}
\begin{lstlisting}[style=artifactjson]
{
  "bids": [
    {
      "reviewer_id": "U_<random_id>",
      "paper_id": "P0042",
      "label": "Very High",
      "score": 2.0,
      "reasoning":
        "Strong match to my expertise."
    }
  ]
}
\end{lstlisting}
\end{minipage}

\vspace{0.65em}

\begin{minipage}[t]{0.485\textwidth}
\textbf{Assignment record: \texttt{assignments.json}}
\begin{lstlisting}[style=artifactjson]
    {
      "metrics": {
      "total_assignments": 300,
      "papers_covered": 100,
      "global_avg_affinity": 0.4749
    }
  "assignments": [
    {
      "paper_id": "P0005",
      "reviewer_id": "U_<random_id>",
      "affinity": 0.4438,
      "bid_score": 2.0,
      "aggregate_score": 4.4438,
      "paper_quality": "strong_accept"
    }
  ]
}
\end{lstlisting}
\end{minipage}
\hfill
\begin{minipage}[t]{0.485\textwidth}
\textbf{Review record: \texttt{review\_scores.json}}
\begin{lstlisting}[style=artifactjson]
{
    "review_stats": {
      "global": {
        "quality_score_correlation": 0.6915,
        "mean_score": 4.42,
        "std_score": 1.79
      }
    }
  "scores": [
    {
      "paper_id": "P0016",
      "reviewer_id": "U_<random_id>",
      "score": 7.0,
      "review_text": "[generated review]",
      "behavior_type": "honest",
      "affinity": 0.6253,
      "quality": "strong_accept"
    }
  ]
}
\end{lstlisting}
\end{minipage}

\caption{Representative world-specific behavioral artifacts. Persona files
encode the experimental behavior condition; bid, assignment, and review files
record successive stages of the simulated conference pipeline. Text fields are
shortened in the illustration, while numerical fields retain their original
types and scales.}
\label{fig:data-schema-worlds}
\end{figure*}

\paragraph{Artifact integrity checks.}
Before computing evaluation metrics, we verify referential consistency across
the artifacts: every paper and reviewer identifier must resolve to the fixed
conference substrate; conflicted paper-reviewer pairs must not appear in the
assignment; every reviewer must produce 20 bids; every paper must receive three
distinct reviewers; and every assigned pair must have exactly one corresponding
review record. Under each behavioral condition, this yields 2,800 bids, 300
assignments, and 300 reviews. Matched worlds are additionally checked to ensure
that they share the same profiles, submissions, affinity scores, and
conflict-of-interest relations.

\paragraph{Responsible release.}
Although profile identifiers are pseudonymous, publication portfolios,
institutional affiliations, coauthor relations, and citation statistics can
serve as quasi-identifiers. We therefore do not plan to publicly release the
unaltered collected-profile snapshot. The public artifact package will retain
the exact file schemas while replacing profile identifiers through a consistent
random mapping and removing profile-derived identifying fields. Synthetic
submissions, experimental configurations, and sanitized numerical bid,
assignment, review, and metric records can be released under the same mapping.
Free-text bid rationales and reviews will be audited to remove inadvertent
references to identifiable profile information. This design supports
recomputation of the reported statistics from sanitized run outputs, while
intentionally separating such reproducibility from redistribution of the
underlying real-world profiles.

\subsection{Run-Level Assignment Access}
\label{sec:appx-rq1-results}
We complement the aggregate assignment-access results with paired within-run
tests. For each run and collusion rate, the binary assignment outcome
$M_{pr}^{(c)}$ for every support relation $(r,p)\in\mathcal{S}$ is paired with
the corresponding outcome $M_{pr}^{(h)}$ in the matched all-honest world.
The collusive condition produces a significant increase in targeted assignment
access in every run under both $r=0.2$ and $r=0.5$ (exact McNemar test,
$p<0.05$). We treat these tests as supplementary paired checks; the primary
evidence is the large assignment-access effect and its consistency across all
seven runs.


\subsection{Target-Paper Score Inflation by Reviewer and Strategy}
\label{sec:appx-rq2-strategies}

Tab.~\ref{tab:appx-rq2-reviewer-scores} decomposes target-paper score
inflation at the reviewer level. Supporting ring members assign substantially
higher scores to captured target papers than honest co-reviewers evaluating
the same papers, yielding within-paper gaps of $+2.21$ and $+1.99$ points under
$r=0.2$ and $r=0.5$, respectively.

The strategy-specific results show that both collusive reviewing policies favor
target papers. Aggressive reviewers assign average scores of $7.45$ and $7.44$,
whereas subtle reviewers assign $6.50$ and $6.27$. The subtle policy therefore
reduces, but does not eliminate, target favoritism. Because strategies are
sampled at the ring level, their realized allocation varies across runs; the
corresponding colluder counts are reported in
Tab.~\ref{tab:appx-strategy-allocation}.

\begin{table}[t]
    \centering
    \small
    \caption{\textbf{Reviewer-level evidence of target-paper score inflation.}
    Values summarize the scores assigned to captured target papers.
    The within-paper gap compares supporting ring members with honest
    co-reviewers evaluating the same papers. Aggressive and subtle columns
    report the corresponding strategy-specific supporter scores.}
    \label{tab:appx-rq2-reviewer-scores}
    \setlength{\tabcolsep}{4.5pt}
    \begin{tabular}{lccccc}
        \toprule
        Scenario
        & \makecell{All\\supporters}
        & \makecell{Honest\\co-reviewers}
        & \makecell{Within-paper\\gap}
        & \makecell{Aggressive\\supporters}
        & \makecell{Subtle\\supporters} \\
        \midrule
        $r=0.2$
        & $7.11$
        & $4.90$
        & $+2.21$
        & $7.45$
        & $6.50$ \\
        $r=0.5$
        & $6.83$
        & $4.83$
        & $+1.99$
        & $7.44$
        & $6.27$ \\
        \bottomrule
    \end{tabular}

    \vspace{2pt}
    \parbox{\linewidth}{%
        \scriptsize
        \emph{Notes.}
        Scores use the shared $1$-$10$ rubric.
        ``All supporters'' combines aggressive and subtle ring reviewers.
        Strategy-specific columns report supporter scores only; the
        within-paper gap is computed for all supporting reviewers against
        honest co-reviewers on the same captured papers.
    }
\end{table}

\subsection{Metrics for Conference-Wide Effects}
\label{sec:appx-rq3-metrics}

Let $N=|\mathcal{P}|=100$. Since each paper receives three reviews, every run
contains $3N=300$ assigned reviewer-paper pairs. The conference-wide mean
review score in world $z$ is
\begin{equation}
    \mu^{(z)}
    =
    \frac{1}{3N}
    \sum_{p\in\mathcal{P}}
    \sum_{j\in\mathcal{R}}
    M_{pj}^{(z)}Y_{pj}^{(z)}.
\end{equation}
The fraction of reviews above the acceptance-level score threshold is
\begin{equation}
    \pi^{(z)}
    =
    \frac{1}{3N}
    \sum_{p\in\mathcal{P}}
    \sum_{j\in\mathcal{R}}
    M_{pj}^{(z)}
    \mathbb{I}\!\left[Y_{pj}^{(z)}>5.5\right].
\end{equation}
This quantity describes the prevalence of favorable individual evaluations; it
does not represent the conference's final paper acceptance rate.

For each paper, we compute the mean of its three assigned reviews as
\begin{equation}
    \overline{Y}_p^{(z)}
    =
    \frac{1}{3}
    \sum_{j\in\mathcal{R}}
    M_{pj}^{(z)}Y_{pj}^{(z)}.
\end{equation}
Quality alignment is then evaluated by the Pearson correlation
\begin{equation}
    \rho_{\mathrm{P}}^{(z)}
    =
    \operatorname{corr}
    \left(
        \{Q_p\}_{p\in\mathcal{P}},
        \{\overline{Y}_p^{(z)}\}_{p\in\mathcal{P}}
    \right)
\end{equation}
and the Spearman rank correlation
\begin{equation}
    \rho_{\mathrm{S}}^{(z)}
    =
    \operatorname{corr}
    \left(
        \{\operatorname{rank}(Q_p)\}_{p\in\mathcal{P}},
        \{\operatorname{rank}(\overline{Y}_p^{(z)})\}_{p\in\mathcal{P}}
    \right).
\end{equation}

Finally, let $\operatorname{TopK}(\mathbf{x})$ denote the indices of the $K$
largest entries in vector $\mathbf{x}$. The top-$30$ overlap is
\begin{equation}
    O_{30}^{(z)}
    =
    \left|
        \operatorname{TopK}_{30}
        \left(\{Q_p\}_{p\in\mathcal{P}}\right)
        \cap
        \operatorname{TopK}_{30}
        \left(\{\overline{Y}_p^{(z)}\}_{p\in\mathcal{P}}\right)
    \right|.
\end{equation}

All five metrics are computed separately for each run and behavioral condition.
Tab.~\ref{tab:global_integrity} reports their mean and standard deviation over
the seven runs.

\section{Detector Evaluation Details}
\label{sec:appx-detectors}

\subsection{Evaluation Goal and Experimental Logic}
\label{sec:appx-Evaluation Goal and Experimental Logic}
    This evaluation asks a deliberately scoped question: given only the bidding
    and authorship information available at the end of the bidding phase, do
    representative bid-based detectors recover the reviewers and rings generated
    by \alg? Our goal is not to establish that \alg is undetectable in general.
    Instead, we test whether its behavior produces the specific malicious-bidding
    signatures assumed by existing detector families.
    
    All detectors operate without access to the realized colluder identities,
    ring memberships, target-paper annotations, reviewer assignments, or review
    scores. Depending on the method, the observable bids are represented in one
    of three ways: as reviewer-level bidding vectors, as a reviewer-reviewer
    bid-author graph, or as a reviewer-paper bipartite graph. These
    representations correspond to different detection hypotheses. Reviewer-level
    methods search for individually anomalous bidding profiles; bid-author graph
    methods search for an unusually dense group of reviewers who bid on one
    another's papers, and the bipartite method searches for a group of reviewers
    whose bids concentrate on a common set of papers. The first family returns a
    ranking of all reviewers, whereas the latter two return one suspicious
    reviewer set. The methods therefore do not form a single interchangeable
    leaderboard and must be evaluated according to their respective outputs.
    
    The common evaluation pipeline is
    [
    \text{bids and authorship}
    $\longrightarrow$
    \text{detector-specific representation}
    $\longrightarrow$
    \text{reviewer ranking or suspicious set}
    $\longrightarrow$
    \text{comparison with the realized collusion structure}
    ]
    Ground-truth colluder and ring labels enter only in the final comparison. For
    each collusive condition, we first measure how well the detector output
    recovers the realized colluders. We then apply the same detector to the
    all-honest reference world and track the reviewers who are designated as
    colluders in the corresponding collusive condition. This second comparison
    asks whether their apparent suspiciousness is induced by collusive behavior
    or is already present in benign affinity-driven bidding and co-authorship
    structure.
    
    The two comparisons provide complementary evidence. Recovery in the
    collusive world characterizes what a detector would flag from the observed
    bids, while the honest-world comparison diagnoses whether that output is
    specific to the behavioral intervention. The honest world is an experimental
    reference rather than an input available to a deployed detector, and the two
    worlds are generated separately rather than forming a bid-by-bid
    counterfactual pair. Accordingly, we interpret the results as a stress test
    of the evaluated detector-representation combinations, not as a claim of
    general or deployment-level undetectability.

\subsection{Common Data and Reference Worlds}
\label{sec:appx-Common Data and Reference Worlds}
    The detector study uses the fixed conference substrate introduced in
    \S\ref{sec:exp_setup}: 140 reviewers and 100 papers spanning four areas
    of computer science, including 92 author-reviewers. Reviewer and paper
    identities, authorship, affinities, conflicts, and each reviewer's top-$20$
    non-conflict bidding pool are shared across behavioral worlds. Because bids
    are submitted only within these pools, the observed reviewer-paper matrix is
    sparse. And the unsubmitted entry is not an observed Neutral bid.
    
    We evaluate one saved triplet comprising an all-honest world and two collusive
    worlds. The $r=0.2$ instance contains 18 colluders in eight rings, while the
    $r=0.5$ instance contains 47 colluders in 20 rings. All rings contain two or
    three reviewers and are constructed from mutual expertise affinity. Colluders
    assign the strongest interest category to eligible papers written by their
    ring partners and otherwise continue to bid by expertise.
    
    The reported detector results characterize this fixed triplet rather than an
    average over the seven lifecycle runs in the main evaluation. Moreover, the
    worlds are generated by separate LLM calls, so non-target bids are not held
    identical. We therefore treat the all-honest world as a structural reference,
    not as a strict bid-level counterfactual.

\subsection{Detector Families and Their Prior Assumptions}
\label{sec:appx-Detector Families and Their Prior Assumptions}
    We organize the eight evaluated detectors by the behavioral structure they
    assume an attack will produce. The reviewer-level methods adapted from
    Jecmen et al.~\cite{jecmen2023dataset} look for anomalous individual bidding
    profiles. The ring-detection methods adapted from Jecmen et
    al.~\cite{jecmen2025detection} instead search for a dense reviewer group or a
    dense reviewer-paper block. Tab.~\ref{tab:detector_families} makes these
    assumptions explicit and states the corresponding question each method poses
    about \alg.

\begin{table*}[t]
    \centering
    \scriptsize
    \caption{\textbf{Detector families and prior assumptions.}
    Each detector operationalizes a different malicious-bidding signature.
    The final column states the diagnostic question evaluated on \alg;
    it is not an additional input to the detector.}
    \label{tab:detector_families}

    \setlength{\tabcolsep}{4pt}
    \renewcommand{\arraystretch}{1.20}

    \begin{tabular}{|
    p{0.155\textwidth}|
    p{0.205\textwidth}|
    p{0.275\textwidth}|
    p{0.275\textwidth}|
    }
        \hline
        \textbf{Detector}
        & \textbf{Input / output}
        & \textbf{Assumed malicious signature}
        & \textbf{Diagnostic question for \alg} \\
        \hline

        \rowcolor{grayblue}
        \multicolumn{4}{|l|}{
            \textbf{Reviewer-level ranking: individual bidding anomalies}
        } \\
        \hline

        \textbf{Counting}\newline
        \texttt{simple\_detect}
        & Ternary reviewer-paper bids; reviewer ranking
        & Many negative bids, with positive bids concentrated on a small target set
        & Does \alg resemble a rejection-style target-seeking attack? \\
        \hline

        \textbf{Pairwise reciprocity}\newline
        \texttt{cluster\_detect}
        & Ternary bids and authorship; reviewer ranking
        & A reviewer pair bids positively on each other's papers while showing limited interest elsewhere
        & Does collusion add reciprocal interest beyond that induced by benign mutual affinity? \\
        \hline

        \textbf{Low-rank residual}\newline
        \texttt{low\_rank\_detect}
        & Ternary reviewer-paper matrix; reviewer ranking
        & A malicious reviewer's bid vector deviates from the matrix's low-rank preference structure
        & Does target-bid inflation create a reviewer-level matrix anomaly? \\
        \hline

        \rowcolor{grayblue}
        \multicolumn{4}{|l|}{
            \textbf{Bid-author graph: anomalously dense reviewer groups}
        } \\
        \hline

        \textbf{OQC-Greedy}
        & Directed bid-author graph; one reviewer set
        & A ring has more internal bid-author edges than expected under a fixed background density
        & Does \alg form a quasi-clique recoverable along a degree-peeling sequence? \\
        \hline

        \textbf{OQC-Local}
        & Directed bid-author graph; one reviewer set
        & The same excess-density structure appears as a local optimum in the graph
        & Can local search isolate a \alg ring from the surrounding affinity graph? \\
        \hline

        \textbf{TellTail}
        & Mutual bid-author graph; one reviewer set
        & Mutual bids are unexpectedly concentrated after accounting for reviewer degree
        & Does \alg reciprocity remain anomalous after controlling for bidding activity? \\
        \hline

        \textbf{Greedy densest subgraph}
        & Directed bid-author graph; one reviewer set
        & Colluders occupy the graph region with the highest average internal connectivity
        & Do the distributed \alg rings form the dominant dense core? \\
        \hline

        \rowcolor{grayblue}
        \multicolumn{4}{|l|}{
            \textbf{Reviewer-paper bipartite graph: coordinated targeting blocks}
        } \\
        \hline

        \textbf{Fraudar}
        & Thresholded reviewer-paper graph; one reviewer-paper block
        & A reviewer group concentrates bids on the same relatively unpopular papers
        & Do \alg bids form a common coordinated-targeting block rather than dispersed small blocks? \\
        \hline
    \end{tabular}
\end{table*}

    Two distinctions are important for interpreting the later results. First,
    despite its implementation name, \texttt{cluster\_detect} scores reviewer
    pairs and converts the maximum pair score into a reviewer ranking; it does not
    return a recovered cluster. Second, Fraudar does not use the bid-author graph
    or paper authorship, but operates directly on the reviewer-paper bipartite
    graph. OQC, TellTail, Greedy Densest Subgraph, and Fraudar each return one
    suspicious set per run, even though \alg distributes colluders across multiple
    small rings. Performance must therefore be read as evidence about the
    particular signature and output structure assumed by each detector, rather
    than as a generic test of whether any evidence of collusion exists.

\subsection{Aligning \alg with Detector Inputs}
\label{sec:appx-aligning}
    The source detectors expect input representations that differ from \alg's
    sparse, five-level bids. We therefore construct a separate input view for each
    detector family. These transformations are part of the evaluated detector
    configuration and are made explicit because they determine which attack signals
    remain visible.
    
    \textbf{Reviewer-paper ternary matrix.}
    For the reviewer-level ranking detectors, we map the five bid levels to the
    ternary representation expected by the source methods:
    \[
    \widetilde{B}_{jp}
    =
    \begin{cases}
        1,  & B_{pj}\in\{\text{High},\text{Very High}\},\\
        0,  & B_{pj}=\text{Neutral},\\
       -1,  & B_{pj}\in\{\text{Low},\text{Very Low}\}.
    \end{cases}
    \]
    We additionally fill unsubmitted entries with $-1$ and set
    authorship-conflict entries to zero. The former is solely a dense-input
    adaptation and should not be interpreted as an observed negative bid.
    Counting, Pairwise Reciprocity, and Low-rank Residual are each applied once to
    the full $140\times100$ reviewer-paper matrix. Pairwise Reciprocity also uses
    authorship information; for reviewer pairs in which either reviewer authors
    multiple papers, we evaluate all combinations of their authored papers and
    retain the maximum pairwise score. Low-rank Residual retains the first three
    singular components and sums the absolute reconstruction residual over each
    reviewer's row.
    
    \textbf{Bid-author graph.}
    For graph-based detectors, we first threshold the original bid matrix as
    \[
    X_{jp}^{(\tau)}
    =
    \mathbb{I}\!\left[B_{jp}\geq\tau\right],
    \qquad
    \tau\in\{1,2\},
    \]
    where $\tau=1$ retains High and Very High bids, while $\tau=2$ retains only
    Very High bids. Let $Y_{pk}=1$ indicate that reviewer $k$ is an author of
    paper $p$. We then construct the directed reviewer graph
    \[
    G_{jk}^{(\tau)}
    =
    \mathbb{I}\!\left[
        \sum_p X_{jp}^{(\tau)}Y_{pk}>0
    \right],
    \qquad
    G_{jj}^{(\tau)}=0.
    \]
    Thus, $j\rightarrow k$ records that reviewer $j$ positively bids on at least
    one paper authored by reviewer $k$; multiple such papers do not increase the
    binary edge weight. The $\tau=1$ view corresponds to the original
    positive-bid semantics, whereas $\tau=2$ provides a diagnostic view of
    \alg's strongest target-bid signal. OQC and Greedy Densest Subgraph operate on
    the directed graph, whereas TellTail retains only mutual edges through
    $\mathbf{G}^{(\tau)}\odot(\mathbf{G}^{(\tau)})^\top$.
    
    \textbf{Reviewer-paper bipartite graph.}
    Fraudar receives the thresholded matrix $\mathbf{X}^{(\tau)}$ directly and
    does not use the authorship projection. Its two node types are reviewers and
    papers. To reduce the influence of papers attracting many positive bids, a
    paper with degree $d_p=\sum_j X_{jp}^{(\tau)}$ receives weight
    $1/\log(d_p+5)$. Fraudar then searches for a dense weighted reviewer-paper
    block and returns the reviewer side of that block for evaluation.
    
    \textbf{Implementation-specific adaptations.}
    Following the source method, we set $\alpha=1/3$ for both OQC variants.
    OQC-Local and TellTail evaluate one heuristic initialization together with ten
    random initializations per invocation. The source Densest Subgraph
    implementation uses an exact linear program requiring a commercial solver; we
    therefore use a greedy degree-peeling approximation and refer to it as Greedy
    Densest Subgraph. Repetition and aggregation of stochastic detector outputs are
    specified in \S\ref{sec:appx-evaluation-measures}.

\subsection{Evaluation Measures and Reporting Conventions}
\label{sec:appx-evaluation-measures}

Let $\mathcal C$ denote all realized colluders in a collusive world and
$\{\mathcal R_g\}$ their ring partition. Because the detector families return
different objects, we evaluate reviewer rankings and suspicious sets
separately.

\textbf{Reviewer-ranking outputs.}
A ranking detector $D$ assigns every reviewer a suspiciousness ordering. We
normalize each zero-based position by the number of reviewers and write it as
$r_D^{(z)}(j)$ for reviewer $j$ in world $z\in\{h,c\}$. Lower values indicate
greater suspiciousness, and the expected mean rank under a random ordering is
approximately $0.5$. For the designated colluders, we report the mean rank in
the honest and collusive worlds,
\[
\bar r_D^{(z)}
=
\frac{1}{|\mathcal C|}
\sum_{j\in\mathcal C}r_D^{(z)}(j),
\qquad
\Delta r_D=\bar r_D^{(c)}-\bar r_D^{(h)}.
\]
Results are displayed as $\bar r_D^{(h)}\!\rightarrow\!\bar r_D^{(c)}$, with
$\Delta r_D$ in parentheses. A negative change means that the designated
colluders move toward the suspicious end under collusion. We also report
Top-$K$ hit,
\[
\operatorname{Hit@K}_D
=
\frac{|\operatorname{TopK}_D^{(c)}\cap\mathcal C|}{|\mathcal C|},
\qquad K=|\mathcal C|,
\]
which measures how many colluders occur among the $K$ most suspicious
reviewers. The raw rank describes absolute prioritization by the detector,
whereas $\Delta r_D$ asks whether collusion creates additional suspiciousness.

\textbf{Set-valued outputs.}
For OQC, TellTail, Greedy Densest Subgraph, and Fraudar, let $S$ be the returned
reviewer set. We report the number of recovered colluders
$q=|S\cap\mathcal C|$ together with the number of flagged reviewers
$m=|S|$. The results tables present this pair directly as $q/m$ (recovered /
flagged), followed by
\[
\operatorname{Precision}(S)=\frac{q}{m},
\qquad
\operatorname{Recall}(S)=\frac{q}{|\mathcal C|}.
\]
Precision measures localization, whereas recall measures coverage of the
distributed collusive population. To determine whether a detector recovers at
least one local ring despite low global recall, we additionally compute
\[
\operatorname{BestRing}(S)
=
\max_g\frac{|S\cap\mathcal R_g|}{|\mathcal R_g|}.
\]
We also compare recall with the expected coverage of a uniformly random set of
the same size, $m/140$; this prevents a broad output from appearing effective
solely because it flags much of the reviewer pool.

\textbf{Honest-world reference.}
Each method is run on the all-honest instance, using the same reviewer IDs in
$\mathcal C$ as a reference group. For a set-valued detector, we display the
mean overlap as
$|S^{(h)}\cap\mathcal C|\!\rightarrow\!|S^{(c)}\cap\mathcal C|$ and define
\[
\Delta_{\mathrm{overlap}}
=
|S^{(c)}\cap\mathcal C|-|S^{(h)}\cap\mathcal C|.
\]
Raw ranks and recovery metrics describe the detector output in the collusive
world. Cross-world changes instead diagnose whether the same reviewers were
already prioritized by benign structure; they are not inputs to the detector
or deployable performance measures.

\textbf{Repetition and aggregation.}
The ranking methods are deterministic and are each applied once to the full
matrix. Set-valued methods are invoked 30 times on each fixed input graph.
OQC-Local and TellTail may return different local optima, so their set sizes
and recovery metrics are averaged over these invocations. OQC-Greedy, Greedy
Densest Subgraph, and Fraudar are deterministic and therefore repeat the same
output. These 30 invocations measure detector-optimization variability, not
variation across independently generated conference instances.

\subsection{Results by Detector Family}
\label{sec:appx-detector-results}

\subsubsection{Reviewer-Level Bidding Anomalies}
\label{sec:appx-ranking-results}

Tab.~\ref{tab:det_family1} reports both the colluders' absolute positions in
the honest and collusive rankings and their change between worlds. Showing the
two ranks is essential: a detector may prioritize the designated reviewers in
the collusive world because they were already unusual under honest bidding,
rather than because their collusive behavior creates a new signal.

\begin{table*}[t]
    \centering
    \small
    \caption{\textbf{Reviewer-level ranking results.}
    For each collusion rate, entries report the designated reviewers' mean
    normalized rank as honest $\rightarrow$ collusive, with $\Delta r$ in
    parentheses. Lower ranks indicate greater suspiciousness, and $0.5$ is the
    random-ranking reference. Top-$K$ hit $(r=.2/.5)$ hit is evaluated in the collusive world
    with $K=|\mathcal C|$.}
    \label{tab:det_family1}

    \setlength{\tabcolsep}{7pt}
    \renewcommand{\arraystretch}{1.20}

    \begin{tabular}{@{}
        p{0.21\textwidth}
        >{\centering\arraybackslash}p{0.27\textwidth}
        >{\centering\arraybackslash}p{0.27\textwidth}
        >{\centering\arraybackslash}p{0.14\textwidth}
        @{}}
        \toprule
        Detector
        & $r=0.2$: Honest $\rightarrow$ Collusive
        & $r=0.5$: Honest $\rightarrow$ Collusive
        & Top-$K$ hit \\
        \midrule

        \rowcolor{grayblue}
        \multicolumn{4}{l}{
            \textbf{Reviewer-level anomaly ranking}
        } \\

        \textbf{Counting}
        & $.551\rightarrow.605\;(+.054)$
        & $.579\rightarrow.604\;(+.025)$
        & $.11/.11$ \\

        \textbf{Pairwise reciprocity}
        & $.292\rightarrow.287\;(-.005)$
        & $.346\rightarrow.332\;(-.014)$
        & $.11/.49$ \\

        \textbf{Low-rank residual}
        & $.463\rightarrow.431\;(-.032)$
        & $.484\rightarrow.460\;(-.024)$
        & $.17/.40$ \\

        \bottomrule
    \end{tabular}
\end{table*}

Counting becomes less suspicious under collusion because its prior is
directionally mismatched to \alg. The method expects malicious reviewers to
reject broadly and reserve positive bids for a few targets, whereas \alg adds
Very High target bids without replacing expertise-based non-target behavior.

Pairwise Reciprocity illustrates why absolute detection scores alone are
insufficient. Under $r=0.2$, the designated reviewers already have a mean rank
of $.292$ when honest and move only to $.287$ when colluding. Likewise, its
Top-$K$ hit reaches $.49$ under $r=0.5$, but the mean-rank change remains only
$-.014$. The detector is finding a real reciprocal structure, but that
structure largely predates the attack because \alg selects rings from mutual
affinity neighborhoods.

Low-rank Residual is the only ranking method with a consistently negative
change, but the shifts are small ($-.032$ and $-.024$) and depend on the dense
encoding of unsubmitted bids. The result supports a weak global perturbation
of the bidding matrix, not clean reviewer-level identification.

\subsubsection{Dense Groups in the Bid-Author Graph}
\label{sec:appx-ring-detectors}

Tab.~\ref{tab:det_family2} separates localization from coverage by showing
the recovered and flagged counts directly. It also shows whether the same
designated reviewers were already selected in the honest-world graph.

\begin{table*}[t]
    \centering
    \scriptsize
    \caption{\textbf{Bid-author graph recovery.}
    ``Rec./flag.'' reports mean recovered colluders $|S\cap\mathcal C|$
    followed by mean flagged reviewers $|S|$. ``Prec./rec.'' reports precision
    and global recall. Honest $\rightarrow$ collusive gives the detector's
    overlap with the designated reviewers in the two worlds. Stochastic
    outputs are averaged over 30 detector invocations on the fixed graph.}
    \label{tab:det_family2}
    \setlength{\tabcolsep}{3.8pt}
    \renewcommand{\arraystretch}{1.16}
    \begin{tabular}{@{}p{0.18\textwidth}cccccc@{}}
        \toprule
        & \multicolumn{3}{c}{$r=0.2$ ($|\mathcal C|=18$)}
        & \multicolumn{3}{c}{$r=0.5$ ($|\mathcal C|=47$)} \\
        \cmidrule(lr){2-4}\cmidrule(l){5-7}
        Detector
        & Rec./flag. & Prec./rec. & H $\rightarrow$ C overlap
        & Rec./flag. & Prec./rec. & H $\rightarrow$ C overlap \\
        \midrule

        \rowcolor{grayblue}
        \multicolumn{7}{l}{\textbf{Native positive-bid graph: $\tau=1$ (High or Very High)}} \\

        OQC-Greedy
        & $2.0/19.0$ & $.105/.111$ & $3.0\rightarrow2.0$
        & $12.0/20.0$ & $.600/.255$ & $14.0\rightarrow12.0$ \\

        OQC-Local
        & $2.0/18.7$ & $.109/.113$ & $2.0\rightarrow2.0$
        & $12.2/20.0$ & $.610/.260$ & $12.3\rightarrow12.2$ \\

        TellTail
        & $2.0/15.0$ & $.133/.111$ & $2.0\rightarrow2.0$
        & $12.3/16.7$ & $.741/.262$ & $12.0\rightarrow12.3$ \\

        Greedy Densest Subgraph
        & $16.0/89.0$ & $.180/.889$ & $17.0\rightarrow16.0$
        & $42.0/77.0$ & $.545/.894$ & $43.0\rightarrow42.0$ \\

        \addlinespace[3pt]
        \rowcolor{grayblue}
        \multicolumn{7}{l}{\textbf{Diagnostic Very-High-only graph: $\tau=2$}} \\

        OQC-Greedy
        & $3.0/8.0$ & $.375/.167$ & $2.0\rightarrow3.0$
        & $6.0/7.0$ & $.857/.128$ & $6.0\rightarrow6.0$ \\

        OQC-Local
        & $1.0/8.0$ & $.125/.056$ & $2.0\rightarrow1.0$
        & $6.0/8.0$ & $.750/.128$ & $5.0\rightarrow6.0$ \\

        TellTail
        & $3.2/4.2$ & $.753/.178$ & $2.0\rightarrow3.2$
        & $5.9/7.4$ & $.874/.126$ & $4.0\rightarrow5.9$ \\

        Greedy Densest Subgraph
        & $8.0/35.0$ & $.229/.444$ & $2.0\rightarrow8.0$
        & $10.0/16.0$ & $.625/.213$ & $12.0\rightarrow10.0$ \\

        \bottomrule
    \end{tabular}
\end{table*}

\textbf{Native positive-bid view.}
At $\tau=1$, OQC-Greedy, OQC-Local, and TellTail return sets of 15-20
reviewers. Under $r=0.2$, each recovers only about two of the 18 colluders.
Under $r=0.5$, their precision rises to $.600$-$.741$, but they recover only
about 12 of 47 colluders. More importantly, their honest-to-collusive overlaps
are nearly unchanged or decrease: for example, OQC-Greedy moves from 14 to 12
designated reviewers. The positive-bid graph therefore contains groups 
that overlap with eventual colluders, but much of that structure is already induced by
honest affinity-based bidding.

Greedy Densest Subgraph displays a different failure mode. It recovers 16 of
18 and 42 of 47 colluders, but does so by flagging 89 and 77 of the 140
reviewers. Its honest-world overlaps are even larger (17 and 43). High recall
here reflects selection of a broad background graph core rather than precise
localization of the planted rings.

\textbf{Diagnostic Very-High-only view.}
At $\tau=2$, removing ordinary High-bid edges exposes smaller structures.
TellTail returns only $4.2$ and $7.4$ reviewers on average, of whom $3.2$ and
$5.9$ are colluders, yielding precision $.753$ and $.874$. It attains
$\operatorname{BestRing}=1$ at both rates, showing that its output contains at
least one complete ring. OQC-Greedy similarly recovers a complete ring, and
at $r=0.5$ identifies six colluders among seven flagged reviewers. These local
successes nevertheless cover only $0.126$-$0.178$ of the full collusive
population. Moreover, high absolute precision does not always represent an
attack-induced change: OQC-Greedy selects six designated reviewers in both
the honest and collusive $r=0.5$ comparisons.

Greedy Densest Subgraph again produces broader outputs. Its overlap increases
from two to eight reviewers under $r=0.2$, but decreases from 12 to 10 under
$r=0.5$. The threshold reveals some local Very-High-bid structure, but does
not yield a consistent global dense core across collusion rates.

\subsubsection{Coordinated Reviewer-Paper Blocks}
\label{sec:appx-fraudar-results}

Fraudar evaluates a different structural hypothesis from the bid-author
methods. Tab.~\ref{tab:det_fraudar} reports whether colluders form a dominant
dense block in the thresholded reviewer-paper graph.

\begin{table*}[t]
    \centering
    \scriptsize
    \caption{\textbf{Fraudar reviewer-paper block recovery.}
    Fraudar returns a reviewer-paper block; the table evaluates its reviewer
    side. Notation follows Tab.~\ref{tab:det_family2}.}
    \label{tab:det_fraudar}

    \setlength{\tabcolsep}{4pt}
    \renewcommand{\arraystretch}{1.20}

    \begin{tabular}{@{}
        p{0.18\textwidth}
        ccc
        ccc
        @{}}
        \toprule
        Input view
        & \multicolumn{3}{c}{$r=0.2$ ($|\mathcal C|=18$)}
        & \multicolumn{3}{c}{$r=0.5$ ($|\mathcal C|=47$)} \\
        \cmidrule(lr){2-4}
        \cmidrule(l){5-7}

        & Rec./flag.
        & Prec./rec.
        & H $\rightarrow$ C overlap
        & Rec./flag.
        & Prec./rec.
        & H $\rightarrow$ C overlap \\
        \midrule

        \rowcolor{grayblue}
        \multicolumn{7}{l}{
            \textbf{Native positive-bid view}
        } \\

        $\tau=1$ (Positive-bids)
        & $15.0/97.0$
        & $.155/.833$
        & $13.0\rightarrow15.0$
        & $41.0/101.0$
        & $.406/.872$
        & $36.0\rightarrow41.0$ \\

        \addlinespace[3pt]

        \rowcolor{grayblue}
        \multicolumn{7}{l}{
            \textbf{Very-High-only diagnostic view}
        } \\

        $\tau=2$ (Very High only)
        & $6.0/36.0$
        & $.167/.333$
        & $2.0\rightarrow6.0$
        & $9.0/21.0$
        & $.429/.191$
        & $4.0\rightarrow9.0$ \\

        \bottomrule
    \end{tabular}
\end{table*}

On the native $\tau=1$ input, Fraudar recovers most colluders but flags 97 and
101 reviewers, corresponding to roughly 69-72\% of the full reviewer pool.
Its recall is therefore high ($.833/.872$), but the output is only about
$1.2\times$ as enriched as a uniformly random set of the same size. As with
the broad densest-subgraph outputs, this is coverage without useful
localization.

The $\tau=2$ view reduces the detected blocks to 36 and 21 reviewers and
increases honest-to-collusive overlap from $2\rightarrow6$ and
$4\rightarrow9$. This indicates that Very High target bids create a visible
bipartite signal. The signal remains incomplete, however: precision is
$.167/.429$ and global recall is $.333/.191$. CABAL's targets are distributed
across many small rings rather than concentrated in one common reviewer-paper
block, so Fraudar does not cleanly isolate the full collusive population.

\subsection{What the Experiment Reveals}
\label{sec:appx-detector-takeaways}

\textbf{Detectability depends on the assumed attack mechanism.}
The Counting result is not evidence that all individual-level evidence is
absent; it shows that a rejection-style prior is directionally mismatched to
\alg's bid-inflation policy. Low-rank Residual detects a small global change
under one dense encoding, while Pairwise Reciprocity detects reviewer pairs
that are already unusual under honest bidding. The three ranking methods fail
for different reasons and should not be summarized as a single negative
result.

\textbf{Benign affinity confounds ordinary positive-bid structure.}
Both Pairwise Reciprocity and the $\tau=1$ bid-author graph expose the main
identification challenge. Because \alg constructs rings from mutual expertise
affinity, its members can legitimately express interest in one another's
papers. Thresholding High and Very High bids into the same binary edge further
removes the distinction between ordinary expertise-based interest and the
strongest target bids. Consequently, a detector can rank or select colluders
without observing a meaningful change when those reviewers become collusive.

\textbf{Very High bids expose a local, representation-dependent signal.}
The $\tau=2$ view removes much of the positive-bid background and allows
TellTail and OQC-Greedy to recover at least one complete small ring with high
precision. Fraudar also exhibits a larger honest-to-collusive overlap on this
view. This finding rules out the stronger claim that \alg leaves no detectable
bidding footprint. However, $\tau=2$ is an attack-informed diagnostic
projection rather than the native positive-bid input of the source methods,
and its signal does not produce high global coverage.

\textbf{Detector output structure limits multi-ring recovery.}
Each graph or bipartite method returns one suspicious block, while \alg
distributes 18 or 47 colluders across eight or 20 rings. Methods that isolate
one local ring therefore achieve high precision but low global recall. Methods
that cover most colluders instead return sets containing a large fraction of
the reviewer pool. The observed precision-coverage trade-off reflects both
the bidding signal and a mismatch between single-block detector outputs and a
distributed multi-ring attack.

Taken together, the results support a narrower conclusion than general
undetectability: \alg leaves fragmented, locally detectable signals, but the
evaluated detector-input combinations do not cleanly separate its distributed
collusive population from benign affinity structure. This conclusion is
limited to the fixed bidding triplet, the input adaptations in
\S\ref{sec:appx-aligning}, and detector-level restarts rather than
independent conference replications. The honest-world comparisons remain
experimental diagnostics, not deployable detector inputs or strictly paired
causal estimates.

\section{Discussion}
\label{sec:appx-discussion}
\textbf{Detection should be pathway-aware.}
Because \alg constructs rings from mutual expertise affinity, colluders target
papers on which they could plausibly express interest under honest behavior.
Positive bids and reciprocal bid-author patterns are therefore not sufficient
evidence of collusion. Consistent with this construction, our detector
comparison shows that native bid graphs confound attack-induced structure with
benign affinity, while the conference-wide score footprint remains modest.
Effective auditing may instead need to condition bidding patterns on expected
expertise and then examine whether suspicious relations convert
disproportionately into reciprocal assignments and systematically divergent
target-paper evaluations. Combining evidence across bidding, assignment, and
reviewing may provide more informative signals than isolated bid anomalies,
although such signals should trigger further investigation rather than serve as
direct evidence of misconduct.

\textbf{Assignment should be treated as an integrity-critical mechanism.}
Our results show that coordinated bids can substantially increase target-paper
access, after which biased reviewing produces the largest downstream effects.
Defenses should therefore act before or during assignment, rather than relying
only on post-hoc review anomalies. Possible directions include limiting the
deterministic influence of bids, introducing controlled randomization,
constraining suspicious reciprocal assignments, and monitoring unusually high
bid-to-assignment conversion within connected reviewer groups. These
interventions must nevertheless preserve legitimate expertise-based bidding
and avoid penalizing tightly connected research communities. \alg provides a
controlled setting for jointly comparing manipulation resistance, assignment
quality, reviewer workload, and false-positive risks across such mechanisms.

\section{Limitation}
\label{sec:appx-limitations}
\alg is designed as a controlled stress-testing framework rather than a faithful
replica of every real-world conference. The current evaluation uses synthetic
submissions, LLM-based reviewer simulacra and reference assessments, and a fixed
conference configuration with a lightweight assignment mechanism. These design
choices enable controlled comparisons across behavioral worlds, but they do not
fully capture the heterogeneity of human reviewers, operational assignment
systems, or broader forms of strategic behavior. Our findings should therefore
be interpreted as evidence about how expertise-grounded collusive bidding can
propagate through the modeled bidding--assignment--reviewing pathway, rather
than as estimates of real-world prevalence or guarantees about detectability in
deployment.

\textbf{Future work.}
Future work will extend \alg by validating its findings across additional model
families, human quality assessments, and richer paper representations;
integrating production-oriented assignment algorithms and a wider range of
conference scales and bidding policies; and studying larger, adaptive, and
multi-stage collusion strategies. Extending the simulation to discussion,
rebuttal, meta-review, and final decision stages would also enable a more
complete account of downstream effects. These directions would strengthen the
framework's external validity and support systematic comparisons of assignment
defenses and collusion-detection methods under more diverse operational
conditions.

\section{Broader Impacts}
\label{sec:appx-broader-impacts}

\textbf{Potential benefits.}
\alg provides a controlled environment for studying how strategic bidding can
affect reviewer assignment and downstream evaluation. It may help conference
organizers identify vulnerabilities before deployment, compare
manipulation-resistant assignment mechanisms, and develop auditing procedures
that improve the integrity and trustworthiness of scientific peer review.
More broadly, the framework supports reproducible research on peer-review
security without requiring experiments on an active conference.

\textbf{Risks and mitigations.}
The same simulations could be misused to refine collusive strategies or search
for behaviors that evade existing detectors. Automated detection also carries
a risk of falsely implicating legitimate reviewers, particularly in small or
closely connected research communities where reciprocal expertise and bidding
patterns arise naturally. Moreover, applying such methods to real bidding or
review data would raise privacy and governance concerns. \alg should therefore
be treated as a defensive stress-testing tool rather than an operational basis
for accusing or sanctioning individuals. Any deployment should protect
confidential conference data, use multiple sources of evidence, audit detector
performance across research communities, and retain human oversight for all
consequential decisions. Where attack artifacts are released, their scope and
level of operational detail should be limited to what is necessary for
reproducible defensive evaluation.


\end{document}